\documentclass{article} 
\usepackage[preprint]{colm2025_conference}

\usepackage{microtype}
\usepackage{hyperref}
\usepackage{url}
\usepackage{booktabs}

\usepackage{lineno}

\usepackage{forest}
\usepackage{amsmath}
\usepackage{wrapfig}
\usepackage{graphicx} 
\usepackage[utf8]{inputenc}
\usepackage{tcolorbox}
\usepackage{amssymb}
\usepackage{fontawesome5}
\definecolor{darkblue}{rgb}{0, 0, 0.5}
\useforestlibrary{edges}
\hypersetup{colorlinks=true, citecolor=darkblue, linkcolor=darkblue, urlcolor=darkblue}

\title{Why Reasoning Matters?\\A Survey of Advancements in Multimodal Reasoning (v1)}

\author{
Jing Bi$^{1}$, Susan Liang$^{1}$, Xiaofei Zhou$^{1}$, Pinxin Liu$^{1}$, Junjia Guo$^{1}$, Yunlong Tang$^{1}$, \And  Luchuan Song$^{1}$, Chao Huang$^{1}$, Ali Vosoughi$^{1}$, Guangyu Sun$^{2}$, Jinxi He$^{1}$, Jiarui Wu$^{1}$, Shu Yang$^{1}$,\And Daoan Zhang$^{1}$, Chen Chen$^{2}$, Lianggong Bruce Wen$^{3}$ , Zhang Liu$^{3}$, Jiebo Luo$^{1\dagger}$, \And Chenliang Xu
 $^{1\dagger}$ \\ \\
$^{1}$University of Rochester, $^{2}$University of Central Florida, $^{3}$Corning Inc.\\ \\
{\tt\small \{jing.bi, sliang22, lsong11, yunlong.tang, ali.vosoughi, daoan.zhang, chenliang.xu\}@rochester.edu} \\
{\tt\small \{jguo40, jhe44, jwu114, syang87\}@ur.rochester.edu} \\
{\tt\small guangyu@ucf.edu, chen.chen@crcv.ucf.edu}\\
{\tt\small \{Wenlb, liuz2\}@corning.com} \\
}

\begin{document}

\ifcolmsubmission
\linenumbers
\fi

\maketitle
\begin{abstract}
Reasoning is central to human intelligence, enabling structured problem-solving across diverse tasks. 
Recent advances in large language models (LLMs) have greatly enhanced their reasoning abilities in arithmetic, commonsense, and symbolic domains.
However, effectively extending these capabilities into multimodal contexts—where models must integrate both visual and textual inputs—continues to be a significant challenge.
Multimodal reasoning introduces complexities, such as handling conflicting information across modalities, which require models to adopt advanced interpretative strategies. 
Addressing these challenges involves not only sophisticated algorithms but also robust methodologies for evaluating reasoning accuracy and coherence.
This paper offers a concise yet insightful overview of reasoning techniques in both textual and multimodal LLMs.
Through a thorough and up-to-date comparison, we clearly formulate core reasoning challenges and opportunities, highlighting practical methods for post-training optimization and test-time inference.
Our work provides valuable insights and guidance, bridging theoretical frameworks and practical implementations, and sets clear directions for future research.
\end{abstract}
\section{Introduction}

Reasoning is a fundamental aspect of human intelligence, enabling us to solve complex problems effectively. 
In LLMs, reasoning has shown promising capabilities, leading to significant advancements in various domains~\citep{Bi_2024}, including arithmetic, commonsense, and symbolic reasoning.
For instance, Chain-of-Thought (CoT) prompting mimics human stepwise problem-solving to boost LLM performance, especially in very large models.

Beyond CoT, other prompting strategies, such as generated knowledge prompting and tree search algorithms (e.g., STAR search~\citep{2203.14465-zelikman_star:_2022}), have been incorporated into LLM training to further enhance reasoning capabilities. 
These methods encourage the model to articulate its internal reasoning in natural language, which reinforces its understanding of the task. 
By laying out intermediate steps, the model can verify and adjust its logic before arriving at the final answer.
\begin{wrapfigure}{rb}{0.44\textwidth}
  \centering
  \includegraphics[width=0.43\textwidth]{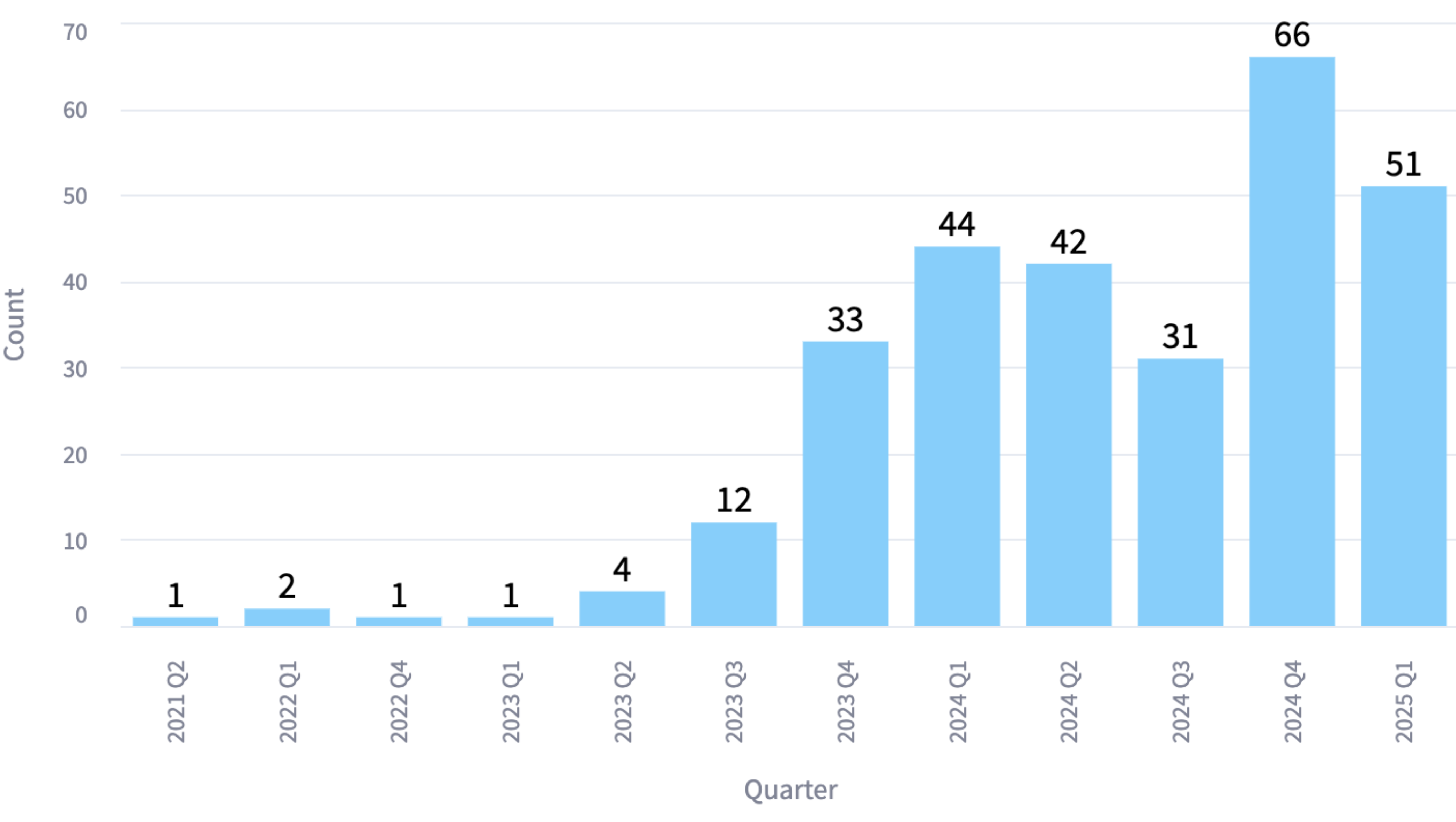}
  \caption{Papers on visual reasoning per quarter over the last three years, with state computed using referenced papers. (Data current to mid-March 2025.)}
  \label{fig:papers}
  \vspace{-8mm}
\end{wrapfigure}

While reasoning in LLMs has progressed significantly, extending these capabilities to multimodal tasks remains an emerging frontier, as multimodal large language models (MLLMs) must navigate the added complexity of interpreting and integrating information from both visual and textual modalities, often resolving ambiguities, inconsistencies, or gaps that arise when the two sources conflict or diverge.

Effective reasoning is central to enabling MLLMs to achieve a compositional understanding—allowing them to deconstruct complex tasks into interpretable, modality-spanning steps. 
It also supports iterative processes such as error correction and self-refinement, where models can revise their outputs by reevaluating both visual and textual cues.

Moreover, strong reasoning capabilities help clear up confusion by inferring spatial relationships, handling counterfactuals, and choosing appropriate tools or actions. These mechanisms also play a crucial role in mitigating hallucinations by grounding model outputs in cross-modal evidence, ultimately improving both accuracy and trustworthiness. Finally, reasoning extends the potential of MLLMs to handle hypothetical scenarios, helping them look ahead and explore “what-if” situations that span both language and vision.

Initial research has demonstrated that integrating Chain-of-Thought (CoT) reasoning across different modalities can significantly enhance the performance of MLLMs. For example,~\cite{2412.18319-yao_mulberry:_2024, 2410.02884-zhang_llama-berry:_2024, bi2024unveilingvisualperceptionlanguage} reports that the LLama series models' reasoning abilities are greatly improved by solely utilizing filtered self-generated reasoning paths.

Building on this momentum, our work provides a comprehensive overview of reasoning in both LLMs and MLLMs, with a focus on its techniques, applications, and future directions. 
We aim to bridge the gap between text-based and multimodal reasoning, offering insights into how reasoning can further enhance the capabilities of LLMs in multimodal contexts.
In the following sections, we begin with a clear problem formulation and definition of reasoning, followed by an in-depth analysis of reasoning techniques—from post-training strategies to test-time computation. We then examine recent trends in datasets and benchmarking. Our goal is to offer a well-organized and accessible survey that supports both theoretical understanding and practical implementation of reasoning in LLMs and MLLMs.

\section{Background}
\label{sec:background}

In complex question-answering tasks, directly predicting an answer can be highly uncertain due to the vast range of possible responses. A more effective approach involves breaking down the reasoning process into a sequence of intermediate steps. This structured method not only improves interpretability but also helps reduce uncertainty at each inference step.

Formally, let $Q$ denote a given question. Conventional language models typically aim to model the conditional probability $P(A \mid Q)$ of the answer $A$ given the question. However, as the complexity of $Q$ increases, the prediction becomes more uncertain—reflected by a rise in the conditional entropy $H(A \mid Q)$. A natural approach to mitigate this uncertainty is to decompose the reasoning process into a sequence of intermediate steps. This leads to modeling the answer generation as a structured chain of conditional probabilities:

\begin{equation}
  P(\text{Step}_1 \mid Q) \cdot P(\text{Step}_2 \mid \text{Step}_1, Q) \cdot \dots \cdot P(\text{Step}_t \mid \text{Step}_{1:t-1}, Q) \cdot P(A \mid \text{Step}_{1:t}, Q)  
\end{equation}

Such a decomposition encourages the model to reduce uncertainty at each stage and improves interpretability and robustness in complex question answering. 
This step decomposition is useful because conditioning on prior steps reduces uncertainty. Specifically, the conditional entropy satisfies: $H(\text{Step}_t \mid \text{Step}_{<t}, Q) \leq H(\text{Step}_t \mid Q)$, meaning each step becomes easier to predict as more context accumulates. From a probabilistic standpoint, the entire reasoning process can be viewed as a trajectory, denoted as $\tau = (\text{Step}_1, \dots, \text{Step}_t)$, which represents a complete reasoning path. Using this, the overall probability of an answer given a question can be written as a marginal over all possible trajectories: 

\begin{equation}
P(A \mid Q) = \sum_{\tau} P(A \mid \tau, Q) \cdot P(\tau \mid Q).
\end{equation}

However, this formulation presents a challenge: enumerating all possible reasoning paths $\tau$ is intractable. 
In practice, CoT reasoning offers a practical workaround by sampling a single trajectory $\hat{\tau}$ from the distribution $P(\tau \mid Q)$, and using it to approximate the answer:

\begin{equation}
P(A \mid Q) \approx P(A \mid \hat{\tau}, Q), \quad \text{where} \quad \hat{\tau} \sim P(\tau \mid Q).
\end{equation}

This is beneficial for many tasks if the model is capable of generating \textbf{one} coherent reasoning path that captures the correct structure of the solution, thereby reducing uncertainty and improving answer accuracy.
More advanced methods such as Tree-of-Thought (ToT) expand on CoT by generating and evaluating multiple reasoning paths, and can be seen as a multiple point estimation approach. In this case, the model generates a set of reasoning paths $\{\tau_1, \tau_2, \ldots, \tau_n\}$ and evaluates them to select the most promising one.

Given the challenge of enumerating all possible reasoning paths, our objective becomes clear: to generate and identify only the most promising trajectories. Achieving this effectively requires two key components:
(i) Generation – a strong model capable of producing high-quality reasoning paths $\tau$;
(ii) Selection – a reliable mechanism for evaluating and choosing the best candidate path.

\begin{figure}
    \centering
    \includegraphics[width=0.9\linewidth]{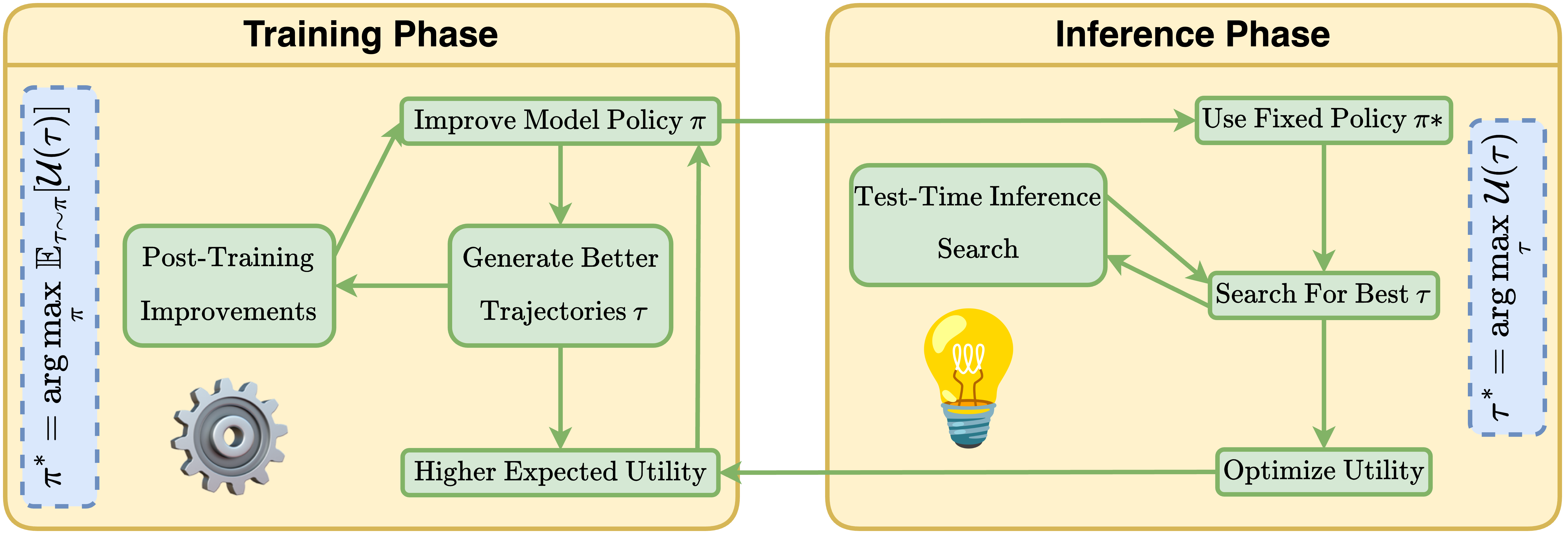}
    \caption{Framework illustrating training and inference for reasoning optimization. A virtuous cycle emerges as better policies generate improved trajectories, which in turn enhance the model through stronger supervision. }
    \label{fig:two}
    \vspace{-5mm}
\end{figure}

These components are deeply interconnected: improved generation policies yield better trajectories, which in turn provide stronger supervision signals to further refine the model. This creates a virtuous cycle of mutual enhancement, as illustrated in Figure~\ref{fig:two}.

At inference time, we want to search for the best $\tau$ under the current model. During training, we aim to improve the model so that it generates reasoning paths that are more likely to lead to correct answers.

Broadly speaking, recent approaches fall into two main categories:

\textbf{Post-training improvement} involves optimizing the model policy $\pi$ to better align outputs with desired utility, typically through techniques like fine-tuning or reinforcement learning. Formally, this corresponds to solving Equation~\eqref{eq:inference}, where the goal is to learn a policy that maximizes expected utility over generated outputs $\tau$.

\textbf{Test-time compute} focuses on improving LLM performance during inference without modifying the model’s core parameters. This aligns with the objective in Equation~\eqref{eq:inference}, where the goal is to select the most useful output trajectory $\tau$ under a fixed policy $\pi^*$, optimizing accuracy, coherence, and efficiency through dynamic strategies like CoT, ToT, and search-based reasoning.

\begin{equation}
\begin{cases}
\pi^* = \displaystyle\arg\max_{\pi} \, \mathbb{E}_{\tau \sim \pi}[\text{Utility}(\tau)] & \text{(Training: Learn a better model)}  \\
\tau^* = \displaystyle\arg\max_{\tau \sim \pi^*} \, \text{Utility}(\tau) & \text{(Inference: Select best trajectory)} 
\label{eq:inference}
\end{cases}
\end{equation}

In the following sections, we will discuss these two components in detail.

\section{Post-training improvements}
\label{sec:post-training}

To enhance the quality of reasoning, we aim to learn a model that generates high-utility reasoning trajectories. One effective approach frames this as a Markov Decision Process (MDP), where the model is trained to maximize the expected return of a reasoning path $\tau$:

\begin{align}
\max_\theta \ \mathbb{E}_{\tau \sim \pi_\theta} \left[ \sum_{t} \gamma^t R(s_t, a_t) \right]
\quad \text{with} \quad
\nabla_\theta J(\pi_\theta) = \mathbb{E}_{\tau}\left[\sum_{t=1}^{T} \nabla_\theta \log \pi_\theta(a_t \mid s_t) A(s_t, a_t)\right]
\end{align}
$s_t$ is the sequence of tokens up to time $t$, and $a_t$ is the next token. The advantage function $A(s_t, a_t)$ estimates the benefit of choosing $a_t$ in state $s_t$, guiding the model toward higher-utility token sequences.
This objective can be directly optimized with respect to the model parameters using policy gradient methods~\citep{williams1992simple}. Notably, this formulation enables the training of LLMs not merely to predict the next token, but to generate complete reasoning trajectories that are optimized for long-term rewards.
\tcbset{
  colback=blue!3!white,
  colframe=blue!30!black,
  boxrule=0.5pt,
  arc=6pt,
  left=4pt,
  right=4pt,
  top=5pt,
  bottom=5pt,
  boxsep=4pt,
}
\begin{tcolorbox}[]
{\Large\textcolor{yellow!70!orange}{\faLightbulb}} \textbf{Policy optimization} is still in its early stages, often relying on existing methods with a primary focus on text token optimization. However, approaches from visual reinforcement learning are worth exploring to better align policies with perceptual and spatial dynamics.
\textbf{Reward alignment} remains heavily reliant on textual signals, highlighting an opportunity to shift focus toward visual-based rewards. Developing methods that can interpret and adapt to visual feedback—such as spatial cues, object dynamics, and scene changes—will be key to achieving more grounded and effective visual decision-making.
In terms of \textbf{model architecture}, improvements often stem from better attention to visual details that align with long-horizon reasoning steps.
\textbf{Spatial-temporal modeling} offers unique opportunities for more creative action definition to manipulate the visual information for better reasoning.
Similarly, during \textbf{data curation}, verifiers commonly depend on grounding models or language cues—indicating potential for more creative, vision-focused verifier designs.
\end{tcolorbox}

\subsection{Policy Optimization}
To solve the above reward maximization problem, recent advancements in visual reasoning have prominently focused on integrating reinforcement learning (RL) and imitation learning (IL) to align MLLMs more closely with human reasoning. IL approaches, such as Thought Cloning~\citep{2503.08525-wei_gtr:_2025}, surpass traditional cloning by aligning intermediate reasoning steps instead of final outputs alone. Coupled with iterative methods like DAgger, this reduces dataset shift and hallucination. LLaVA-Critic~\citep{2410.02712-xiong_llava-critic:_2024} further generalizes reward signals via preference-based alignment, boosting multimodal reasoning effectiveness with minimal modality-specific fine-tuning. RL from Simulations (RLS3)~\citep{2501.18880-waite_rls3:_2025}, employing Soft Actor-Critic (SAC) agents, significantly enhances spatial reasoning accuracy and efficiency in models like PaliGemma, outperforming random exploration baselines~\citep{2501.18880-waite_rls3:_2025}. FuRL~\citep{2406.00645-fu_furl:_2024} addresses RL reward misalignment through dual-stage alignment, effectively mitigating sparse reward challenges. Adaptive Reward Programming (ARP-DT+)~\citep{2309.10790-kim_guide_2023}, integrating CLIP representations with Decision Transformers, enhances reasoning robustness across domains by employing active inference to counter distractions and domain shifts.

\subsection{Reward Alignment}
Recent approaches have explored iterative refinement and reflection mechanisms, token-level rewards, and automated benchmarking to enhance reasoning quality and mitigate hallucinations. 
PARM++~\citep{2501.13926-guo_can_2025} introduces a reflection-based mechanism that iteratively refines outputs by identifying misalignments and actively correcting them through self-assessment loops, significantly improving test-time GenEval scores up to three reflection cycles~\citep{2501.13926-guo_can_2025}. Similarly, REVERIE employs rationale-based training, explicitly supervising reasoning steps, thus improving coherence and reducing hallucinations compared to models without rationales~\citep{2407.11422-zhang_reflective_2024}.
Fine-grained reward models, such as MMViG~\citep{2402.06118-yan_vigor:_2024} and Reward Alignment via Preference Learning (RAPL)~\citep{2310.07932-tian_what_2024}, emphasize granular feedback at individual reasoning steps, achieving precise improvements without manual reward tuning. Similarly, CLIP-DPO~\citep{2408.10433-ouali_clip-dpo:_2024} leverages pre-trained CLIP directly as a reward function, streamlining reward modeling.
Challenges such as constrained perceptual fields are addressed through dynamic evaluation strategies like zooming and shifting, emphasizing the importance of sequential decision-making in visual reasoning~\citep{2410.04659-wang_actiview:_2024}. Structural issues like positional bias and patch-boundary effects further underscore the need for targeted architectural optimizations~\citep{2402.07384-zhang_exploring_2024}.
Recent approaches integrate reward signals directly into generation. TLDR~\citep{2410.04734-fu_tldr:_2024} applies token-level binary rewards, enabling real-time feedback to mitigate hallucinations and boost both annotation efficiency and backbone model performance~\citep{2410.04734-fu_tldr:_2024}. Complementarily, EACO enhances intermediate object recognition, significantly reducing hallucinations compared to earlier models like LLaVA-v1.6-7B~\citep{2412.04903-wang_eaco:_2024}.
Distillation frameworks also contribute: FIRE leverages iterative student–teacher feedback and structured evaluation to refine visual reasoning~\citep{2407.11522-li_fire:_2024}, while SILKIE demonstrates the scalability of GPT-4V-generated feedback in improving perception and cognition without human intervention~\citep{2312.10665-li_silkie:_2023}.

\subsection{Model Architecture}
Nested architectures, such as MaGNeTS, have emerged as effective means of balancing computational efficiency with model accuracy by employing parameter-sharing and caching mechanisms during inference~\citep{2502.00382-goyal_masked_2025}.  Intrinsic activation approaches, notably ROSS, bypass external modules by integrating multimodal understanding directly into the model’s core, yielding improved adaptability to diverse visual inputs, including depth maps~\citep{2410.09575-wang_reconstructive_2024}. Additionally, DIFFLMM incorporates diffusion models and attention-based segmentation, enhancing visual grounding precision~\citep{2410.08209-cao_emerging_2024}, while Mini-Monkey targets the "sawtooth effect" in lightweight models through adaptive cropping and scale compression, achieving superior visual-text alignment without extensive computational resources~\citep{2408.02034-huang_mini-monkey:_2024}. SEA dynamically adjusts embedding alignment strategies to maintain performance across varying resolutions and model sizes~\citep{2408.11813-yin_sea:_2024}. Concurrently, PAE-LaMM demonstrates the combined effectiveness of vision encoder adaptation and pixel reconstruction tasks, systematically enhancing visual detail awareness and question-answering performance~\citep{2408.03940-gou_how_2024}.
Conversely, LLAVIDAL and EventLens uniquely integrate domain-specific visual elements, such as object-centric embeddings and event-aware tokens, respectively, directly within LLM architectures to effectively reason about activities and temporal contexts~\citep{2406.09390-reilly_llavidal:_2024,2404.13847-ma_eventlens:_2024}.

\subsection{Spatial Temporal Modeling}
Unlike pure language reasoning, visual reasoning involves a richer action space, allowing operations such as zooming, cropping, and frame selection, which enables the model to interact with and manipulate visual inputs. The GeoGLIP pipeline utilizes geometric pre-training with a dynamic feature router to advance fine-grained visual mathematical understanding by balancing symbolic and visual modalities based on task demands~\citep{2501.06430-zhang_open_2025}. Concurrently, hallucination issues are mitigated via targeted fine-tuning and prompting strategies grounded in recent vision-language work like CapQA~\citep{2501.02964-hu_socratic_2025}.
For spatial reasoning, simpler graph topologies have been shown to enhance performance, while structural complexity reduces accuracy in edge-centric tasks~\citep{2412.13540-zhu_benchmarking_2024}. Augmentation via uncertainty-aware active inference further boosts attribute detection precision~\citep{2412.07012-zhang_provision:_2024}.
Intrinsic spatial-temporal modeling is advanced through architectures embedding multi-modal understanding without reliance on external depth tools~\citep{2410.09575-wang_reconstructive_2024}, and dual-branch structures improve video temporal grounding in multi-hop reasoning~\citep{2408.14469-chen_grounded_2024}.
The shift toward multimodal benchmarks better reflects real-world spatial-temporal reasoning~\citep{2408.08632-li_survey_2024}. Integrating rationalization with answer generation enhances temporal reasoning~\citep{2407.06189-zohar_video-star:_2024}, while automated annotation improves data scalability and bridges gaps between commercial and open-source Video-LLMs~\citep{2406.11303-li_videovista:_2024}.
Multi-modal integration and active inference optimize modality selection for multi-step spatial-temporal reasoning~\citep{2405.03272-zhang_worldqa:_2024}. Adjusting learning objectives to minimize hallucinations improves precision and EOS decision-making~\citep{2402.14545-yue_less_2024}.

\begin{figure}[h]
    \centering
    \includegraphics[width=1\linewidth]{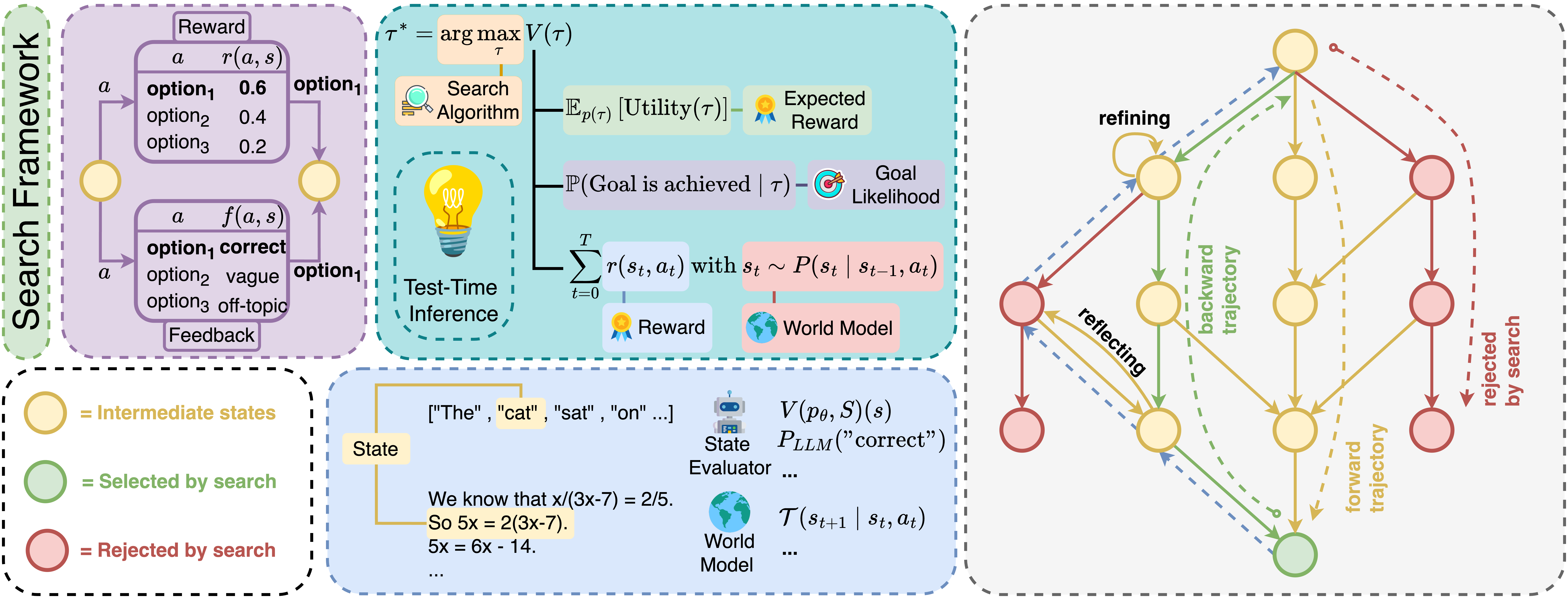}
    \caption{Search framework where language models explore and refine reasoning paths. Trajectories are scored using reward models, based on expected utility or final output quality, and guided by feedback, world models, and evaluators to select the most promising steps.}
    \label{fig:search}
    \vspace{-3mm}
\end{figure}
\section{Test-Time compute}

At test time, the goal is to find the reasoning trajectory $\tau$ that maximizes a utility function as shown as red and green path in Figure~\ref{fig:search}

\begin{equation}
    \tau^* = \arg\max_{\tau} \, \mathcal{U}(\tau), \quad \text{where } \mathcal{U}(\tau) =
\begin{cases}
\sum_{t} r(s_t, a_t) & \text{(MDP-style reward)} \\
\mathbb{P}(\text{goal} \mid \tau) & \text{(Goal likelihood)} \\
\mathbb{P}(\tau \succ \tau') & \text{(Preference-based)} \\
f(\text{rank}(\tau)) & \text{(Ranking-based)} \\
\vdots & \text{(Others: risk-sensitive, etc.)}
\end{cases}
\end{equation}

The utility $\mathcal{U}(\tau)$ can be defined in various ways, including cumulative rewards (MDP-style), goal likelihood, preference comparisons, and ranking-based scores. The choice of utility depends on the supervision available and the nature of the task. The core challenge remains sampling high-utility trajectories under the chosen formulation.

Since directly modeling $\mathcal{U}(\tau)$ is often difficult, we use reward models to approximate it. These models can include (1) absolute rewards (as in standard RL), (2) pairwise preference models (e.g., DPO), and (3) ranking-based models.

\tcbset{
  colback=green!3!white,
  colframe=green!30!black,
  boxrule=0.5pt,
  arc=6pt,
  left=4pt,
  right=4pt,
  top=5pt,
  bottom=5pt,
  boxsep=4pt,
}
\begin{tcolorbox}[]
{\Large\textcolor{yellow!70!orange}{\faLightbulb}} \textbf{Search Strategies} Current search algorithms primarily operate on distinct textual tokens. However, innovative approaches are emerging for raw visual tokens, termed Creative Visual Search, expanding search capabilities directly within visual spaces. Moreover, Multi-Granularity Spatial Search is gaining attention as an effective extension, enhancing exploration across diverse spatial granularities.
\textbf{Reward and Feedback} Reward systems can significantly benefit from deeper visual insights. Leveraging visual prompts encourages models to engage in self-reflection grounded in visual information, substantially improving semantic alignment and decision-making effectiveness.
\textbf{Iterative Refinement} Determining the alignment between textual tokens and visual information remains challenging. Consequently, model self-reflection can produce hallucinated interpretations. Nevertheless, these reflections offer valuable insights, highlighting new pathways to mitigate unnecessary or hallucinated reflections and enhance model reliability.
\end{tcolorbox}
\subsection{Search Strategies}

Monte Carlo Tree Search (MCTS) has emerged as a powerful tool for managing uncertainty and sequential decision-making. \citet{2501.10674-imam_can_2025} applied MCTS to enhance temporal reasoning, reducing hallucinations in time-dependent data. Building on this, CoMCTS\citep{2412.18319-yao_mulberry:_2024} unified multiple MLLMs within a collaborative tree search, addressing model bias and improving accuracy and efficiency. Vision-specific strategies advanced with VisVM~\citep{2412.03704-wang_scaling_2024}, which estimates long-term candidate value to reduce hallucinations, outperforming immediate-alignment methods like CLIP-PRM. Video reasoning similarly benefited from reward modeling and adaptive exploration to prioritize key frames~\citep{2412.10471-yang_vca:_2024}. Structured search methods also gained traction. LLaVA-o1~\citep{2411.10440-xu_llava-cot:_2025} uses stage-level beam search for complex reasoning, while MVP~\citep{2408.17150-qu_look_2024} aggregates certainty across multi-perspective captions to resist adversarial inputs. DC2~\citep{2408.15556-wang_divide_2024} applies MCTS-based cropping to focus on salient image regions for high-resolution reasoning.
Multimodal and temporal search frameworks like VideoVista~\citep{2406.11303-li_videovista:_2024}, WorldRetriver~\citep{2405.03272-zhang_worldqa:_2024}, and DynRefer~\citep{2405.16071-zhao_dynrefer:_2024} surpass static baselines using adaptive sampling, fusion, and stochastic inference. Step-by-step comparative reasoning with LMs also enhances video QA~\citep{2404.16222-nagarajan_step_2024}. 
Innovations in context refinement (VURF~\citep{2403.14743-mahmood_vurf:_2024}), image decomposition (V*\citep{2312.14135-wu_v*:_2023}), and dynamic tool use (AVIS\citep{2306.08129-hu_avis:_2023}) highlight the shift toward adaptive visual reasoning. FuRL~\citep{2406.00645-fu_furl:_2024} aligns reward modeling with iterative fine-tuning for better performance.
In embodied AI, combining chain-of-thought, self-verification, and MCTS-driven planning enables scalable, robust decision-making in dynamic environments~\citep{2404.15190-shin_socratic_2024}.

\subsection{Adaptive Inference}
Adaptive inference is reshaping vision-language reasoning by enabling dynamic, context-sensitive processing that improves both accuracy and efficiency. Central to this is the iterative evaluation and refinement of outputs, often through internal feedback and external verification.
LOOKBACK enhances correction accuracy through iterative visual re-examination within each reasoning step~\citep{2412.02172-wu_visco:_2024}. IXC-2.5 balances performance and response length via Best-of-N sampling guided by reward models~\citep{2501.12368-zang_internlm-xcomposer2.5-reward:_2025}, while PARM++ uses reflection-based refinement to align generated images with prompts~\citep{2501.13926-guo_can_2025}.
Further innovations include ProgressCaptioner’s sliding-window approach for tracking action progress over time~\citep{2412.02071-xue_progress-aware_2024}.
To combat multi-modal hallucinations, MEMVR triggers visual retracing based on uncertainty, optimizing cost-accuracy tradeoffs~\citep{2410.03577-zou_look_2024}, while MVP aggregates certainty across diverse views~\citep{2408.17150-qu_look_2024}. SID applies token-level contrastive decoding to filter irrelevant content~\citep{2408.02032-huo_self-introspective_2024}.
Models like DualFocus integrate macro and micro reasoning for fine-grained attention control~\citep{2402.14767-cao_dualfocus:_2024}, and LISA++ enables test-time compute scaling without retraining~\citep{2312.17240-yang_lisa++:_2024}. PerceptionGPT accelerates inference via adaptive token embeddings that encode spatial cues and dynamically weigh layers~\citep{2311.06612-pi_perceptiongpt:_2023}.

\subsection{Reward}
\citet{2501.13926-guo_can_2025} introduce ORM and PRM, showing that ORM’s final-step evaluation significantly enhances image generation through ranking data. PARM++ further refines this method by enabling iterative reflection for self-correction, thus improving prompt fidelity. Complementarily, \citet{2501.05444-hao_can_2025} highlight persistent errors in VLM spatial reasoning and explanation, proposing reward re-ranking as an interim solution.
To assess VL-GenRMs more comprehensively, \citet{2411.17451-li_vlrewardbench:_2024} propose VL-RewardBench, pinpointing perception and reasoning challenges while advocating critic training and co-evolutionary reward learning. Self-training methods also gain prominence: R3V~\citep{2411.00855-cheng_vision-language_2024} leverages synthesized rationales to enhance noise-robust Chain-of-Thought (CoT) refinement, while CECE~\citep{2410.22315-cascante-bonilla_natural_2024} enriches evaluation using LLM-generated entailment/contradiction captions.
Enhanced supervision methods also emerge. \citet{2410.16198-zhang_improve_2024} enhance CoT reasoning through ShareGPT-4O distillation, Supervised Fine-Tuning (SFT), and Direct Preference Optimization (DPO). LLaVA-Critic~\citep{2410.02712-xiong_llava-critic:_2024} builds upon this using critic-guided DPO without external human feedback. FuRL~\citep{2406.00645-fu_furl:_2024} introduces a two-stage tuning approach addressing misalignments and temporal errors, while HYDRA~\citep{2403.12884-ke_hydra:_2024} employs dynamic loops for adaptive instruction ranking. Additionally, m\&m’s~\citep{2403.11085-ma_m&ms:_2024} provides a structured framework for evaluating multi-step planning via tool and argument accuracy.

\subsection{Feedback}
VOLCANO employs a three-stage iterative framework—initial response generation, visual self-assessment, and revision—to reduce hallucinations by emphasizing accurate image details~\citep{2311.07362-lee_volcano:_2024}. Similarly, LOOKBACK mandates explicit atomic verification against images to enhance both critique and correction processes for LVLMs~\citep{2412.02172-wu_visco:_2024}. LLaVA-Critic advances feedback by incorporating diverse critic instructions and iterative DPO training, leveraging internal data~\citep{2410.02712-xiong_llava-critic:_2024}. Additionally, tailored visual prompting with iterative binary verification effectively enhances semantic grounding in models like LLaVA-1.5, ViP-LLaVA, and CogVLM~\citep{2404.06510-liao_can_2024}.
Minimalist reinforcement learning setups have shown effectiveness, particularly for multimodal mathematical reasoning, with larger models like InternVL2.5-38B benefiting from difficulty-based data filtering to stabilize training~\citep{2503.07365-meng_mm-eureka:_2025}. Explicit visual CoT mechanisms, exemplified by VisCoT, integrate bounding box predictions and iterative cropping to enhance visual question-answering performance~\citep{2403.16999-shao_visual_2024}. In video understanding, VIDEOTREE utilizes adaptive hierarchical clustering for efficient navigation through structured tree-based representations~\citep{2405.19209-wang_videotree:_2024}. Lastly, prompt engineering strategies like MiHO and MiHI significantly reduce hallucinations without model retraining, post-processing evaluations with GPT-4 further enhancing reliability without architectural modifications~\citep{2402.01345-han_skip_2024,2501.02964-hu_socratic_2025}.

\subsection{Iterative Refinement}
Adaptive inference transforms vision-language reasoning by facilitating dynamic, context-sensitive processing to enhance accuracy and efficiency. Central to this approach is iterative output refinement through internal feedback and external verification. For instance, IXC-2.5 employs Best-of-N sampling guided by reward models to optimize performance and response length~\citep{2501.12368-zang_internlm-xcomposer2.5-reward:_2025}, while PARM++ uses reflection-based refinement for better alignment between images and prompts~\citep{2501.13926-guo_can_2025}.
Further advancements include LLaVA-o1’s structured perception and stage-level beam search for decomposing complex reasoning tasks~\citep{2411.10440-xu_llava-cot:_2025}. To mitigate multi-modal hallucinations, MEMVR implements visual retracing based on uncertainty to optimize the cost-accuracy tradeoff~\citep{2410.03577-zou_look_2024}, MVP aggregates certainty across diverse views~\citep{2408.17150-qu_look_2024}, and SID employs token-level contrastive decoding to eliminate irrelevant content~\citep{2408.02032-huo_self-introspective_2024}.
Moreover, DualFocus integrates macro- and micro-level reasoning for precise attention control~\citep{2402.14767-cao_dualfocus:_2024}, LISA++ enables test-time compute scaling without retraining~\citep{2312.17240-yang_lisa++:_2024}, and PerceptionGPT accelerates inference through adaptive token embeddings encoding spatial cues and dynamically weighted layers~\citep{2311.06612-pi_perceptiongpt:_2023}.
\subsection{Dataset Curation}

Recent work shows that high-quality, strategically curated datasets outperform large but noisy alternatives in guiding reasoning paths. R-CoT~\citep{2410.17885-deng_r-cot:_2024} enhances geometric reasoning via reverse generation and stepwise synthesis. Task-specific datasets such as SMIR~\citep{2501.03675-li_smir:_2025} show up to 8\% performance gains over generic datasets, demonstrating the value of targeted curation. Gradual complexity in dataset construction, as shown in VIREO~\citep{2406.19934-cheng_least_2024}, ensures foundational reasoning skills before introducing advanced tasks. Scalable dataset creation through automation also proves effective. VideoVista~\citep{2406.11303-li_videovista:_2024} uses auto-annotations for video reasoning, outperforming manually curated sets. DecoVQA+~\citep{2409.19339-zhang_visual_2024} explicitly teaches when to apply question decomposition, aided by SelectiveVQD loss for better strategy selection. LocVLM~\citep{2404.07449-ranasinghe_learning_2024} scales with pseudo-data and implicit feedback signals like location and relevance prediction. CogCoM~\citep{2402.04236-qi_cogcom:_2024} incorporates tasks like grounding and manipulation to reinforce step-by-step reasoning.

\section{Datasets and Benchmarks}

\noindent \textbf{Structured and Task-Specific Reasoning}
Datasets such as \textit{Visual-RFT}\citep{2503.01785-liu_visual-rft:_2025}, \textit{CapQA}\citep{2501.02964-hu_socratic_2025}, \textit{GUIDE}\citep{2406.18227-liang_guide:_2024}, \textit{STAR}\citep{2405.09711-wu_star:_2024}, \textit{Visual Genome}\citep{2407.11422-zhang_reflective_2024}, \textit{VL-GPT}\citep{2312.09251-zhu_vl-gpt:_2023}, and \textit{Interfacing}~\citep{2312.07532-zou_interfacing_2024} have introduced structured prompting that significantly enhances models’ reasoning accuracy by explicitly guiding reasoning processes. In contrast to earlier benchmarks like \textit{STAR}, these datasets place a stronger emphasis on task-specific reward functions and structured inference, aiming to improve generalization across diverse visual and linguistic contexts.

\noindent \textbf{Temporal and Spatial Reasoning}
Temporal benchmarks like \textit{VisualQA, and TemporalVQA}~\citep{2501.10674-imam_can_2025}, \textit{TLQA}~\citep{2501.07214-swetha_timelogic:_2025}, \textit{REXTIME}~\citep{2406.19392-chen_rextime:_2024}, \textit{FrameCap}~\citep{2412.02071-xue_progress-aware_2024}, and \textit{VideoVista}~\citep{2406.11303-li_videovista:_2024} have revealed substantial limitations in current multimodal language models, achieving accuracy significantly below human levels (15\% versus 90\%). These benchmarks emphasize the necessity for temporal logic annotations and highlight the difficulty models face when reasoning across video frames. In parallel, spatial datasets such as \textit{SpatialVLM}~\citep{2401.12168-chen_spatialvlm:_2024}, \textit{WhatsUp}~\citep{2310.19785-kamath_whats_2023}, \textit{DC2}~\citep{2408.15556-wang_divide_2024}, and \textit{Grounded}~\citep{2408.14469-chen_grounded_2024} showcase improvements through extensive spatial reasoning QA pairs and challenging spatial configurations.

\noindent \textbf{Iterative and Reflective Reasoning}
Iterative reasoning capabilities have markedly improved through datasets like \textit{VISCO}~\citep{2412.02172-wu_visco:_2024}, \textit{Mulberry-260k}~\citep{2412.18319-yao_mulberry:_2024}, \textit{FIRE}~\citep{2407.11522-li_fire:_2024}, \textit{VLFeedback}~\citep{2410.09421-li_vlfeedback:_2024}, \textit{Silkie}~\citep{2312.10665-li_silkie:_2023}, \textit{TIIL}~\citep{2404.18033-huang_exposing_2024}, \textit{ConMe}~\citep{2406.08164-huang_conme:_2024}, and \textit{Reflective}~\citep{2407.11422-zhang_reflective_2024}, enabling models to learn from mistakes by both incorporating reflective and iterative feedback loops. For instance, VISCO demonstrated that human critiques significantly outperform model critiques (76\% error correction), highlighting a persistent gap in models' ability to independently self-correct.

\noindent \textbf{Complex Multimodal Evaluation}
Complex multimodal datasets like \textit{EMMA}~\citep{2501.05444-hao_can_2025}, \textit{CoMT}~\citep{2412.12932-cheng_comt:_2024}, \textit{SMIR}~\citep{2501.03675-li_smir:_2025}, \textit{ProVision}~\citep{2412.07012-zhang_provision:_2024}, \textit{MAGEBench}~\citep{2412.04531-zhang_magebench:_2024}, \textit{MM-Vet v2}~\citep{2408.00765-yu_mm-vet_2024}, \textit{JourneyBench}~\citep{2409.12953-wang_journeybench:_2025}, VERIFY~\citep{bi2025verify} and \textit{CompCap}~\citep{2412.05243-chen_compcap:_2024} exposed limitations in current multimodal models' integration capabilities. EMMA specifically shows that models struggle significantly with iterative multimodal interactions, particularly when tasks demand deep integration across domain-specific problems.
Benchmarks like \textit{FineCops-Ref}~\citep{2409.14750-liu_finecops-ref:_2025}, \textit{VL-RewardBench}~\citep{2411.17451-li_vlrewardbench:_2024}, \textit{MM-SAP}~\citep{2401.07529-wang_mm-sap:_2024}, \textit{PCA-Bench}~\citep{2402.15527-chen_pca-bench:_2024}, \textit{AttCoSeg}~\citep{2312.12423-pramanick_jack_2024}, and \textit{M3CoT}~\citep{2405.16473-chen_m$^3$cot:_2024} emphasize detailed evaluation metrics (Recall@k, AUROC). These benchmarks provide insights into models' actual reasoning capabilities beyond traditional accuracy metrics.

\noindent \textbf{Counterfactual and Logical Reasoning}
Datasets like \textit{CounterCurate}~\citep{2402.13254-zhang_countercurate:_2024}, \textit{C-VQA}~\citep{2310.06627-zhang_what_2024}, \textit{CausalChaos!}~\citep{2404.01299-parmar_causalchaos!_2024}, \textit{LogicAI}~\citep{2407.04973-xiao_logicvista:_2024}, and \textit{MCDGRAPH}~\citep{2412.13540-zhu_benchmarking_2024} explicitly test model reasoning robustness, showing improvements through challenging counterfactual examples and structured logical questions.

\noindent \textbf{Active Perception and Progressive Reasoning}
Active perception datasets such as \textit{ActiView}~\citep{2410.04659-wang_actiview:_2024} and progressive reasoning benchmarks such as \textit{Blink}~\citep{2404.12390-fu_blink:_2024}, \textit{ADL-X}~\citep{2406.09390-reilly_llavidal:_2024}, and \textit{GUIDE}~\citep{2406.18227-liang_guide:_2024} challenge models' dynamic reasoning capabilities through active view adjustments and frame-level progression.

\section{Conclusion}
In this paper, we presented a comprehensive survey of existing work on reasoning in Multimodal Large Language Models (MLLMs) and Large Language Models (LLMs), with a focus on optimizing model performance and identifying the most effective reasoning trajectories. We introduced a variety of post-training and test-time computation methods, discussing their potential to enhance model capabilities on complex, real-world tasks. Our analysis highlights the critical role of reasoning in improving the visual understanding abilities of MLLMs. Furthermore, we have provided insights and outlined potential directions for future research to guide the development of more robust and efficient reasoning frameworks.

\bibliography{references}

@incollection{sweller2011cognitive,
  title={Cognitive load theory},
  author={Sweller, John},
  booktitle={Psychology of learning and motivation},
  volume={55},
  pages={37--76},
  year={2011},
  publisher={Elsevier}
}

@article{paas2004cognitive,
  title={Cognitive load theory: Instructional implications of the interaction between information structures and cognitive architecture},
  author={Paas, Fred and Renkl, Alexander and Sweller, John},
  journal={Instructional science},
  volume={32},
  number={1/2},
  pages={1--8},
  year={2004},
  publisher={JSTOR}
}

@book{vygotsky1978mind,
  title={Mind in society: The development of higher psychological processes},
  author={Vygotsky, Lev S},
  volume={86},
  year={1978},
  publisher={Harvard university press}
}

@article{van2010scaffolding,
  title={Scaffolding in teacher--student interaction: A decade of research},
  author={Van de Pol, Janneke and Volman, Monique and Beishuizen, Jos},
  journal={Educational psychology review},
  volume={22},
  pages={271--296},
  year={2010},
  publisher={Springer}
}

@article{lai2011metacognition,
  title={Metacognition: A literature review},
  author={Lai, Emily R},
  year={2011},
  publisher={Pearson Research Report. Pearson Education, Upper Saddle River}
}

@article{frankish2010dual,
  title={Dual-process and dual-system theories of reasoning},
  author={Frankish, Keith},
  journal={Philosophy Compass},
  volume={5},
  number={10},
  pages={914--926},
  year={2010},
  publisher={Wiley Online Library}
}

@incollection{waite2022jean,
  title={Jean Piaget's constructivist theory of learning},
  author={Waite-Stupiansky, Sandra},
  booktitle={Theories of early childhood education},
  pages={3--18},
  year={2022},
  publisher={Routledge}
}

@article{bisra2018inducing,
  title={Inducing self-explanation: A meta-analysis},
  author={Bisra, Kiran and Liu, Qing and Nesbit, John C and Salimi, Farimah and Winne, Philip H},
  journal={Educational Psychology Review},
  volume={30},
  pages={703--725},
  year={2018},
  publisher={Springer}
}

@article{vanlehn1992model,
  title={A model of the self-explanation effect},
  author={VanLehn, Kurt and Jones, Randolph M and Chi, Michelene TH},
  journal={The journal of the learning sciences},
  volume={2},
  number={1},
  pages={1--59},
  year={1992},
  publisher={Taylor \& Francis}
}

@article{chi1994eliciting,
  title={Eliciting self-explanations improves understanding},
  author={Chi, Michelene TH and De Leeuw, Nicholas and Chiu, Mei-Hung and LaVancher, Christian},
  journal={Cognitive science},
  volume={18},
  number={3},
  pages={439--477},
  year={1994},
  publisher={Elsevier}
}

@book{bruner2009process,
  title={The process of education},
  author={Bruner, Jerome S},
  year={2009},
  publisher={Harvard university press}
}

@incollection{gentner2017analogical,
  title={Analogical reasoning},
  author={Gentner, Dedre and Maravilla, Francisco},
  booktitle={International handbook of thinking and reasoning},
  pages={186--203},
  year={2017},
  publisher={Routledge}
}

@misc{2203.14465-zelikman_star:_2022,
 author = {Zelikman, Eric and Wu, Yuhuai and Mu, Jesse and Goodman, Noah D.},
 doi = {10.48550/arXiv.2203.14465},
 month = {May},
 note = {arXiv:2203.14465},
 publisher = {arXiv},
 shorttitle = {{STaR}},
 title = {{STaR}: {Bootstrapping} {Reasoning} {With} {Reasoning}},
 url = {http://arxiv.org/abs/2203.14465},
 urldate = {2025-02-07},
 year = {2022}
}

@misc{2306.08129-hu_avis:_2023,
 author = {Hu, Ziniu and Iscen, Ahmet and Sun, Chen and Chang, Kai-Wei and Sun, Yizhou and Ross, David A. and Schmid, Cordelia and Fathi, Alireza},
 doi = {10.48550/arXiv.2306.08129},
 month = {November},
 note = {arXiv:2306.08129},
 publisher = {arXiv},
 shorttitle = {{AVIS}},
 title = {{AVIS}: {Autonomous} {Visual} {Information} {Seeking} with {Large} {Language} {Model} {Agent}},
 url = {http://arxiv.org/abs/2306.08129},
 urldate = {2025-02-07},
 year = {2023}
}

@misc{2309.10790-kim_guide_2023,
 author = {Kim, Changyeon and Seo, Younggyo and Liu, Hao and Lee, Lisa and Shin, Jinwoo and Lee, Honglak and Lee, Kimin},
 doi = {10.48550/arXiv.2309.10790},
 month = {October},
 note = {arXiv:2309.10790},
 publisher = {arXiv},
 title = {Guide {Your} {Agent} with {Adaptive} {Multimodal} {Rewards}},
 url = {http://arxiv.org/abs/2309.10790},
 urldate = {2025-02-07},
 year = {2023}
}

@misc{2309.15112-zhang_internlm-xcomposer:_2023,
 author = {Zhang, Pan and Dong, Xiaoyi and Wang, Bin and Cao, Yuhang and Xu, Chao and Ouyang, Linke and Zhao, Zhiyuan and Duan, Haodong and Zhang, Songyang and Ding, Shuangrui and Zhang, Wenwei and Yan, Hang and Zhang, Xinyue and Li, Wei and Li, Jingwen and Chen, Kai and He, Conghui and Zhang, Xingcheng and Qiao, Yu and Lin, Dahua and Wang, Jiaqi},
 doi = {10.48550/arXiv.2309.15112},
 month = {December},
 note = {arXiv:2309.15112},
 publisher = {arXiv},
 shorttitle = {{InternLM}-{XComposer}},
 title = {{InternLM}-{XComposer}: {A} {Vision}-{Language} {Large} {Model} for {Advanced} {Text}-image {Comprehension} and {Composition}},
 url = {http://arxiv.org/abs/2309.15112},
 urldate = {2025-02-07},
 year = {2023}
}

@misc{2310.06627-zhang_what_2024,
 author = {Zhang, Letian and Zhai, Xiaotong and Zhao, Zhongkai and Zong, Yongshuo and Wen, Xin and Zhao, Bingchen},
 doi = {10.48550/arXiv.2310.06627},
 month = {April},
 note = {arXiv:2310.06627},
 publisher = {arXiv},
 shorttitle = {What {If} the {TV} {Was} {Off}?},
 title = {What {If} the {TV} {Was} {Off}? {Examining} {Counterfactual} {Reasoning} {Abilities} of {Multi}-modal {Language} {Models}},
 url = {http://arxiv.org/abs/2310.06627},
 urldate = {2025-02-07},
 year = {2024}
}

@misc{2310.07932-tian_what_2024,
 author = {Tian, Ran and Xu, Chenfeng and Tomizuka, Masayoshi and Malik, Jitendra and Bajcsy, Andrea},
 doi = {10.48550/arXiv.2310.07932},
 month = {January},
 note = {arXiv:2310.07932},
 publisher = {arXiv},
 shorttitle = {What {Matters} to {You}?},
 title = {What {Matters} to {You}? {Towards} {Visual} {Representation} {Alignment} for {Robot} {Learning}},
 url = {http://arxiv.org/abs/2310.07932},
 urldate = {2025-02-07},
 year = {2024}
}

@misc{2310.19785-kamath_whats_2023,
 author = {Kamath, Amita and Hessel, Jack and Chang, Kai-Wei},
 doi = {10.48550/arXiv.2310.19785},
 month = {October},
 note = {arXiv:2310.19785},
 publisher = {arXiv},
 shorttitle = {What's "up" with vision-language models?},
 title = {What's "up" with vision-language models? {Investigating} their struggle with spatial reasoning},
 url = {http://arxiv.org/abs/2310.19785},
 urldate = {2025-02-07},
 year = {2023}
}

@misc{2311.03079-wang_cogvlm:_2024,
 author = {Wang, Weihan and Lv, Qingsong and Yu, Wenmeng and Hong, Wenyi and Qi, Ji and Wang, Yan and Ji, Junhui and Yang, Zhuoyi and Zhao, Lei and Song, Xixuan and Xu, Jiazheng and Xu, Bin and Li, Juanzi and Dong, Yuxiao and Ding, Ming and Tang, Jie},
 doi = {10.48550/arXiv.2311.03079},
 month = {February},
 note = {arXiv:2311.03079},
 publisher = {arXiv},
 shorttitle = {{CogVLM}},
 title = {{CogVLM}: {Visual} {Expert} for {Pretrained} {Language} {Models}},
 url = {http://arxiv.org/abs/2311.03079},
 urldate = {2025-02-07},
 year = {2024}
}

@misc{2311.06612-pi_perceptiongpt:_2023,
 author = {Pi, Renjie and Yao, Lewei and Gao, Jiahui and Zhang, Jipeng and Zhang, Tong},
 doi = {10.48550/arXiv.2311.06612},
 month = {November},
 note = {arXiv:2311.06612},
 publisher = {arXiv},
 shorttitle = {{PerceptionGPT}},
 title = {{PerceptionGPT}: {Effectively} {Fusing} {Visual} {Perception} into {LLM}},
 url = {http://arxiv.org/abs/2311.06612},
 urldate = {2025-02-07},
 year = {2023}
}

@misc{2311.07362-lee_volcano:_2024,
 author = {Lee, Seongyun and Park, Sue Hyun and Jo, Yongrae and Seo, Minjoon},
 doi = {10.48550/arXiv.2311.07362},
 month = {April},
 note = {arXiv:2311.07362},
 publisher = {arXiv},
 shorttitle = {Volcano},
 title = {Volcano: {Mitigating} {Multimodal} {Hallucination} through {Self}-{Feedback} {Guided} {Revision}},
 url = {http://arxiv.org/abs/2311.07362},
 urldate = {2025-02-07},
 year = {2024}
}

@misc{2311.13601-li_visual_2023,
 author = {Li, Feng and Jiang, Qing and Zhang, Hao and Ren, Tianhe and Liu, Shilong and Zou, Xueyan and Xu, Huaizhe and Li, Hongyang and Li, Chunyuan and Yang, Jianwei and Zhang, Lei and Gao, Jianfeng},
 doi = {10.48550/arXiv.2311.13601},
 month = {November},
 note = {arXiv:2311.13601},
 publisher = {arXiv},
 title = {Visual {In}-{Context} {Prompting}},
 url = {http://arxiv.org/abs/2311.13601},
 urldate = {2025-02-07},
 year = {2023}
}

@misc{2312.03052-hu_visual_2024,
 author = {Hu, Yushi and Stretcu, Otilia and Lu, Chun-Ta and Viswanathan, Krishnamurthy and Hata, Kenji and Luo, Enming and Krishna, Ranjay and Fuxman, Ariel},
 doi = {10.48550/arXiv.2312.03052},
 month = {April},
 note = {arXiv:2312.03052},
 publisher = {arXiv},
 shorttitle = {Visual {Program} {Distillation}},
 title = {Visual {Program} {Distillation}: {Distilling} {Tools} and {Programmatic} {Reasoning} into {Vision}-{Language} {Models}},
 url = {http://arxiv.org/abs/2312.03052},
 urldate = {2025-02-07},
 year = {2024}
}

@misc{2312.03631-ben-kish_mitigating_2024,
 author = {Ben-Kish, Assaf and Yanuka, Moran and Alper, Morris and Giryes, Raja and Averbuch-Elor, Hadar},
 doi = {10.48550/arXiv.2312.03631},
 month = {October},
 note = {arXiv:2312.03631},
 publisher = {arXiv},
 title = {Mitigating {Open}-{Vocabulary} {Caption} {Hallucinations}},
 url = {http://arxiv.org/abs/2312.03631},
 urldate = {2025-02-07},
 year = {2024}
}

@misc{2312.07532-zou_interfacing_2024,
 author = {Zou, Xueyan and Li, Linjie and Wang, Jianfeng and Yang, Jianwei and Ding, Mingyu and Wei, Junyi and Yang, Zhengyuan and Li, Feng and Zhang, Hao and Liu, Shilong and Aravinthan, Arul and Lee, Yong Jae and Wang, Lijuan},
 doi = {10.48550/arXiv.2312.07532},
 month = {July},
 note = {arXiv:2312.07532},
 publisher = {arXiv},
 title = {Interfacing {Foundation} {Models}' {Embeddings}},
 url = {http://arxiv.org/abs/2312.07532},
 urldate = {2025-02-07},
 year = {2024}
}

@misc{2312.08870-ma_vista-llama:_2023,
 author = {Ma, Fan and Jin, Xiaojie and Wang, Heng and Xian, Yuchen and Feng, Jiashi and Yang, Yi},
 doi = {10.48550/arXiv.2312.08870},
 month = {December},
 note = {arXiv:2312.08870},
 publisher = {arXiv},
 shorttitle = {Vista-{LLaMA}},
 title = {Vista-{LLaMA}: {Reliable} {Video} {Narrator} via {Equal} {Distance} to {Visual} {Tokens}},
 url = {http://arxiv.org/abs/2312.08870},
 urldate = {2025-02-07},
 year = {2023}
}

@misc{2312.09251-zhu_vl-gpt:_2023,
 author = {Zhu, Jinguo and Ding, Xiaohan and Ge, Yixiao and Ge, Yuying and Zhao, Sijie and Zhao, Hengshuang and Wang, Xiaohua and Shan, Ying},
 doi = {10.48550/arXiv.2312.09251},
 month = {December},
 note = {arXiv:2312.09251},
 publisher = {arXiv},
 shorttitle = {{VL}-{GPT}},
 title = {{VL}-{GPT}: {A} {Generative} {Pre}-trained {Transformer} for {Vision} and {Language} {Understanding} and {Generation}},
 url = {http://arxiv.org/abs/2312.09251},
 urldate = {2025-02-07},
 year = {2023}
}

@misc{2312.10665-li_silkie:_2023,
 author = {Li, Lei and Xie, Zhihui and Li, Mukai and Chen, Shunian and Wang, Peiyi and Chen, Liang and Yang, Yazheng and Wang, Benyou and Kong, Lingpeng},
 doi = {10.48550/arXiv.2312.10665},
 month = {December},
 note = {arXiv:2312.10665},
 publisher = {arXiv},
 shorttitle = {Silkie},
 title = {Silkie: {Preference} {Distillation} for {Large} {Visual} {Language} {Models}},
 url = {http://arxiv.org/abs/2312.10665},
 urldate = {2025-02-07},
 year = {2023}
}

@misc{2312.11420-zhao_tuning_2023,
 author = {Zhao, Bingchen and Tu, Haoqin and Wei, Chen and Mei, Jieru and Xie, Cihang},
 doi = {10.48550/arXiv.2312.11420},
 month = {December},
 note = {arXiv:2312.11420},
 publisher = {arXiv},
 shorttitle = {Tuning {LayerNorm} in {Attention}},
 title = {Tuning {LayerNorm} in {Attention}: {Towards} {Efficient} {Multi}-{Modal} {LLM} {Finetuning}},
 url = {http://arxiv.org/abs/2312.11420},
 urldate = {2025-02-07},
 year = {2023}
}

@misc{2312.12423-pramanick_jack_2024,
 author = {Pramanick, Shraman and Han, Guangxing and Hou, Rui and Nag, Sayan and Lim, Ser-Nam and Ballas, Nicolas and Wang, Qifan and Chellappa, Rama and Almahairi, Amjad},
 doi = {10.48550/arXiv.2312.12423},
 month = {June},
 note = {arXiv:2312.12423},
 publisher = {arXiv},
 shorttitle = {Jack of {All} {Tasks}, {Master} of {Many}},
 title = {Jack of {All} {Tasks}, {Master} of {Many}: {Designing} {General}-purpose {Coarse}-to-{Fine} {Vision}-{Language} {Model}},
 url = {http://arxiv.org/abs/2312.12423},
 urldate = {2025-02-07},
 year = {2024}
}

@misc{2312.14135-wu_v*:_2023,
 author = {Wu, Penghao and Xie, Saining},
 doi = {10.48550/arXiv.2312.14135},
 month = {December},
 note = {arXiv:2312.14135},
 publisher = {arXiv},
 shorttitle = {V*},
 title = {V*: {Guided} {Visual} {Search} as a {Core} {Mechanism} in {Multimodal} {LLMs}},
 url = {http://arxiv.org/abs/2312.14135},
 urldate = {2025-02-07},
 year = {2023}
}

@misc{2312.17240-yang_lisa++:_2024,
 author = {Yang, Senqiao and Qu, Tianyuan and Lai, Xin and Tian, Zhuotao and Peng, Bohao and Liu, Shu and Jia, Jiaya},
 doi = {10.48550/arXiv.2312.17240},
 month = {January},
 note = {arXiv:2312.17240},
 publisher = {arXiv},
 shorttitle = {{LISA}++},
 title = {{LISA}++: {An} {Improved} {Baseline} for {Reasoning} {Segmentation} with {Large} {Language} {Model}},
 url = {http://arxiv.org/abs/2312.17240},
 urldate = {2025-02-07},
 year = {2024}
}

@misc{2401.03105-he_incorporating_2024,
 author = {He, Xin and Wei, Longhui and Xie, Lingxi and Tian, Qi},
 doi = {10.48550/arXiv.2401.03105},
 month = {January},
 note = {arXiv:2401.03105},
 publisher = {arXiv},
 title = {Incorporating {Visual} {Experts} to {Resolve} the {Information} {Loss} in {Multimodal} {Large} {Language} {Models}},
 url = {http://arxiv.org/abs/2401.03105},
 urldate = {2025-02-07},
 year = {2024}
}

@misc{2401.07529-wang_mm-sap:_2024,
 author = {Wang, Yuhao and Liao, Yusheng and Liu, Heyang and Liu, Hongcheng and Wang, Yu and Wang, Yanfeng},
 doi = {10.48550/arXiv.2401.07529},
 month = {June},
 note = {arXiv:2401.07529},
 publisher = {arXiv},
 shorttitle = {{MM}-{SAP}},
 title = {{MM}-{SAP}: {A} {Comprehensive} {Benchmark} for {Assessing} {Self}-{Awareness} of {Multimodal} {Large} {Language} {Models} in {Perception}},
 url = {http://arxiv.org/abs/2401.07529},
 urldate = {2025-02-07},
 year = {2024}
}

@misc{2401.12168-chen_spatialvlm:_2024,
 author = {Chen, Boyuan and Xu, Zhuo and Kirmani, Sean and Ichter, Brian and Driess, Danny and Florence, Pete and Sadigh, Dorsa and Guibas, Leonidas and Xia, Fei},
 doi = {10.48550/arXiv.2401.12168},
 month = {January},
 note = {arXiv:2401.12168},
 publisher = {arXiv},
 shorttitle = {{SpatialVLM}},
 title = {{SpatialVLM}: {Endowing} {Vision}-{Language} {Models} with {Spatial} {Reasoning} {Capabilities}},
 url = {http://arxiv.org/abs/2401.12168},
 urldate = {2025-02-07},
 year = {2024}
}

@misc{2401.17981-jiao_training-free_2024,
 author = {Jiao, Qirui and Chen, Daoyuan and Huang, Yilun and Li, Yaliang and Shen, Ying},
 doi = {10.48550/arXiv.2401.17981},
 month = {December},
 note = {arXiv:2401.17981},
 publisher = {arXiv},
 shorttitle = {From {Training}-{Free} to {Adaptive}},
 title = {From {Training}-{Free} to {Adaptive}: {Empirical} {Insights} into {MLLMs}' {Understanding} of {Detection} {Information}},
 url = {http://arxiv.org/abs/2401.17981},
 urldate = {2025-02-07},
 year = {2024}
}

@misc{2402.01345-han_skip_2024,
 author = {Han, Zongbo and Bai, Zechen and Mei, Haiyang and Xu, Qianli and Zhang, Changqing and Shou, Mike Zheng},
 doi = {10.48550/arXiv.2402.01345},
 month = {May},
 note = {arXiv:2402.01345},
 publisher = {arXiv},
 shorttitle = {Skip {\textbackslash}n},
 title = {Skip {\textbackslash}n: {A} {Simple} {Method} to {Reduce} {Hallucination} in {Large} {Vision}-{Language} {Models}},
 url = {http://arxiv.org/abs/2402.01345},
 urldate = {2025-02-07},
 year = {2024}
}

@misc{2402.04236-qi_cogcom:_2024,
 author = {Qi, Ji and Ding, Ming and Wang, Weihan and Bai, Yushi and Lv, Qingsong and Hong, Wenyi and Xu, Bin and Hou, Lei and Li, Juanzi and Dong, Yuxiao and Tang, Jie},
 doi = {10.48550/arXiv.2402.04236},
 month = {May},
 note = {arXiv:2402.04236},
 publisher = {arXiv},
 shorttitle = {{CogCoM}},
 title = {{CogCoM}: {Train} {Large} {Vision}-{Language} {Models} {Diving} into {Details} through {Chain} of {Manipulations}},
 url = {http://arxiv.org/abs/2402.04236},
 urldate = {2025-02-07},
 year = {2024}
}

@misc{2402.06118-yan_vigor:_2024,
 author = {Yan, Siming and Bai, Min and Chen, Weifeng and Zhou, Xiong and Huang, Qixing and Li, Li Erran},
 doi = {10.48550/arXiv.2402.06118},
 month = {October},
 note = {arXiv:2402.06118},
 publisher = {arXiv},
 shorttitle = {{ViGoR}},
 title = {{ViGoR}: {Improving} {Visual} {Grounding} of {Large} {Vision} {Language} {Models} with {Fine}-{Grained} {Reward} {Modeling}},
 url = {http://arxiv.org/abs/2402.06118},
 urldate = {2025-02-07},
 year = {2024}
}

@misc{2402.07384-zhang_exploring_2024,
 author = {Zhang, Jiarui and Hu, Jinyi and Khayatkhoei, Mahyar and Ilievski, Filip and Sun, Maosong},
 doi = {10.48550/arXiv.2402.07384},
 month = {February},
 note = {arXiv:2402.07384},
 publisher = {arXiv},
 title = {Exploring {Perceptual} {Limitation} of {Multimodal} {Large} {Language} {Models}},
 url = {http://arxiv.org/abs/2402.07384},
 urldate = {2025-02-07},
 year = {2024}
}

@misc{2402.13254-zhang_countercurate:_2024,
 author = {Zhang, Jianrui and Cai, Mu and Xie, Tengyang and Lee, Yong Jae},
 doi = {10.48550/arXiv.2402.13254},
 month = {June},
 note = {arXiv:2402.13254},
 publisher = {arXiv},
 shorttitle = {{CounterCurate}},
 title = {{CounterCurate}: {Enhancing} {Physical} and {Semantic} {Visio}-{Linguistic} {Compositional} {Reasoning} via {Counterfactual} {Examples}},
 url = {http://arxiv.org/abs/2402.13254},
 urldate = {2025-02-07},
 year = {2024}
}

@misc{2402.14545-yue_less_2024,
 author = {Yue, Zihao and Zhang, Liang and Jin, Qin},
 doi = {10.48550/arXiv.2402.14545},
 month = {May},
 note = {arXiv:2402.14545},
 publisher = {arXiv},
 shorttitle = {Less is {More}},
 title = {Less is {More}: {Mitigating} {Multimodal} {Hallucination} from an {EOS} {Decision} {Perspective}},
 url = {http://arxiv.org/abs/2402.14545},
 urldate = {2025-02-07},
 year = {2024}
}

@misc{2402.14767-cao_dualfocus:_2024,
 author = {Cao, Yuhang and Zhang, Pan and Dong, Xiaoyi and Lin, Dahua and Wang, Jiaqi},
 doi = {10.48550/arXiv.2402.14767},
 month = {February},
 note = {arXiv:2402.14767},
 publisher = {arXiv},
 shorttitle = {{DualFocus}},
 title = {{DualFocus}: {Integrating} {Macro} and {Micro} {Perspectives} in {Multi}-modal {Large} {Language} {Models}},
 url = {http://arxiv.org/abs/2402.14767},
 urldate = {2025-02-07},
 year = {2024}
}

@misc{2402.15527-chen_pca-bench:_2024,
 author = {Chen, Liang and Zhang, Yichi and Ren, Shuhuai and Zhao, Haozhe and Cai, Zefan and Wang, Yuchi and Wang, Peiyi and Meng, Xiangdi and Liu, Tianyu and Chang, Baobao},
 doi = {10.48550/arXiv.2402.15527},
 month = {February},
 note = {arXiv:2402.15527},
 publisher = {arXiv},
 shorttitle = {{PCA}-{Bench}},
 title = {{PCA}-{Bench}: {Evaluating} {Multimodal} {Large} {Language} {Models} in {Perception}-{Cognition}-{Action} {Chain}},
 url = {http://arxiv.org/abs/2402.15527},
 urldate = {2025-02-07},
 year = {2024}
}

@misc{2403.07487-zhang_motion_2024,
 author = {Zhang, Zeyu and Liu, Akide and Reid, Ian and Hartley, Richard and Zhuang, Bohan and Tang, Hao},
 doi = {10.48550/arXiv.2403.07487},
 month = {August},
 note = {arXiv:2403.07487},
 publisher = {arXiv},
 shorttitle = {Motion {Mamba}},
 title = {Motion {Mamba}: {Efficient} and {Long} {Sequence} {Motion} {Generation}},
 url = {http://arxiv.org/abs/2403.07487},
 urldate = {2025-02-07},
 year = {2024}
}

@misc{2403.09333-zhan_griffon_2024,
 author = {Zhan, Yufei and Zhu, Yousong and Zhao, Hongyin and Yang, Fan and Tang, Ming and Wang, Jinqiao},
 doi = {10.48550/arXiv.2403.09333},
 month = {March},
 note = {arXiv:2403.09333},
 publisher = {arXiv},
 shorttitle = {Griffon v2},
 title = {Griffon v2: {Advancing} {Multimodal} {Perception} with {High}-{Resolution} {Scaling} and {Visual}-{Language} {Co}-{Referring}},
 url = {http://arxiv.org/abs/2403.09333},
 urldate = {2025-02-07},
 year = {2024}
}

@misc{2403.09394-wang_git:_2024,
 author = {Wang, Haiyang and Tang, Hao and Jiang, Li and Shi, Shaoshuai and Naeem, Muhammad Ferjad and Li, Hongsheng and Schiele, Bernt and Wang, Liwei},
 doi = {10.48550/arXiv.2403.09394},
 month = {March},
 note = {arXiv:2403.09394},
 publisher = {arXiv},
 shorttitle = {{GiT}},
 title = {{GiT}: {Towards} {Generalist} {Vision} {Transformer} through {Universal} {Language} {Interface}},
 url = {http://arxiv.org/abs/2403.09394},
 urldate = {2025-02-07},
 year = {2024}
}

@misc{2403.11085-ma_m&ms:_2024,
 author = {Ma, Zixian and Huang, Weikai and Zhang, Jieyu and Gupta, Tanmay and Krishna, Ranjay},
 doi = {10.48550/arXiv.2403.11085},
 month = {September},
 note = {arXiv:2403.11085},
 publisher = {arXiv},
 shorttitle = {m\&m's},
 title = {m\&m's: {A} {Benchmark} to {Evaluate} {Tool}-{Use} for multi-step multi-modal {Tasks}},
 url = {http://arxiv.org/abs/2403.11085},
 urldate = {2025-02-07},
 year = {2024}
}

@misc{2403.12884-ke_hydra:_2024,
 author = {Ke, Fucai and Cai, Zhixi and Jahangard, Simindokht and Wang, Weiqing and Haghighi, Pari Delir and Rezatofighi, Hamid},
 doi = {10.48550/arXiv.2403.12884},
 month = {July},
 note = {arXiv:2403.12884},
 publisher = {arXiv},
 shorttitle = {{HYDRA}},
 title = {{HYDRA}: {A} {Hyper} {Agent} for {Dynamic} {Compositional} {Visual} {Reasoning}},
 url = {http://arxiv.org/abs/2403.12884},
 urldate = {2025-02-07},
 year = {2024}
}

@misc{2403.14743-mahmood_vurf:_2024,
 author = {Mahmood, Ahmad and Vayani, Ashmal and Naseer, Muzammal and Khan, Salman and Khan, Fahad Shahbaz},
 doi = {10.48550/arXiv.2403.14743},
 month = {March},
 note = {arXiv:2403.14743},
 publisher = {arXiv},
 shorttitle = {{VURF}},
 title = {{VURF}: {A} {General}-purpose {Reasoning} and {Self}-refinement {Framework} for {Video} {Understanding}},
 url = {http://arxiv.org/abs/2403.14743},
 urldate = {2025-02-07},
 year = {2024}
}

@misc{2403.16999-shao_visual_2024,
 author = {Shao, Hao and Qian, Shengju and Xiao, Han and Song, Guanglu and Zong, Zhuofan and Wang, Letian and Liu, Yu and Li, Hongsheng},
 doi = {10.48550/arXiv.2403.16999},
 month = {November},
 note = {arXiv:2403.16999},
 publisher = {arXiv},
 shorttitle = {Visual {CoT}},
 title = {Visual {CoT}: {Advancing} {Multi}-{Modal} {Language} {Models} with a {Comprehensive} {Dataset} and {Benchmark} for {Chain}-of-{Thought} {Reasoning}},
 url = {http://arxiv.org/abs/2403.16999},
 urldate = {2025-02-07},
 year = {2024}
}

@misc{2403.19322-chen_plug-and-play_2024,
 author = {Chen, Jiaxing and Liu, Yuxuan and Li, Dehu and An, Xiang and Deng, Weimo and Feng, Ziyong and Zhao, Yongle and Xie, Yin},
 doi = {10.48550/arXiv.2403.19322},
 month = {June},
 note = {arXiv:2403.19322},
 publisher = {arXiv},
 title = {Plug-and-{Play} {Grounding} of {Reasoning} in {Multimodal} {Large} {Language} {Models}},
 url = {http://arxiv.org/abs/2403.19322},
 urldate = {2025-02-07},
 year = {2024}
}

@misc{2404.01258-zhang_direct_2024,
 author = {Zhang, Ruohong and Gui, Liangke and Sun, Zhiqing and Feng, Yihao and Xu, Keyang and Zhang, Yuanhan and Fu, Di and Li, Chunyuan and Hauptmann, Alexander and Bisk, Yonatan and Yang, Yiming},
 doi = {10.48550/arXiv.2404.01258},
 month = {April},
 note = {arXiv:2404.01258},
 publisher = {arXiv},
 title = {Direct {Preference} {Optimization} of {Video} {Large} {Multimodal} {Models} from {Language} {Model} {Reward}},
 url = {http://arxiv.org/abs/2404.01258},
 urldate = {2025-02-07},
 year = {2024}
}

@misc{2404.01299-parmar_causalchaos!_2024,
 author = {Parmar, Paritosh and Peh, Eric and Chen, Ruirui and Lam, Ting En and Chen, Yuhan and Tan, Elston and Fernando, Basura},
 doi = {10.48550/arXiv.2404.01299},
 month = {June},
 note = {arXiv:2404.01299},
 publisher = {arXiv},
 title = {{CausalChaos}! {Dataset} for {Comprehensive} {Causal} {Action} {Question} {Answering} {Over} {Longer} {Causal} {Chains} {Grounded} in {Dynamic} {Visual} {Scenes}},
 url = {http://arxiv.org/abs/2404.01299},
 urldate = {2025-02-07},
 year = {2024}
}

@misc{2404.01911-dzabraev_vlrm:_2024,
 author = {Dzabraev, Maksim and Kunitsyn, Alexander and Ivaniuta, Andrei},
 doi = {10.48550/arXiv.2404.01911},
 month = {April},
 note = {arXiv:2404.01911},
 publisher = {arXiv},
 shorttitle = {{VLRM}},
 title = {{VLRM}: {Vision}-{Language} {Models} act as {Reward} {Models} for {Image} {Captioning}},
 url = {http://arxiv.org/abs/2404.01911},
 urldate = {2025-02-07},
 year = {2024}
}

@misc{2404.04007-liang_neural-symbolic_2024,
 author = {Liang, Lili and Sun, Guanglu and Qiu, Jin and Zhang, Lizhong},
 doi = {10.48550/arXiv.2404.04007},
 month = {April},
 note = {arXiv:2404.04007},
 publisher = {arXiv},
 shorttitle = {Neural-{Symbolic} {VideoQA}},
 title = {Neural-{Symbolic} {VideoQA}: {Learning} {Compositional} {Spatio}-{Temporal} {Reasoning} for {Real}-world {Video} {Question} {Answering}},
 url = {http://arxiv.org/abs/2404.04007},
 urldate = {2025-02-07},
 year = {2024}
}

@misc{2404.06510-liao_can_2024,
 author = {Liao, Yuan-Hong and Mahmood, Rafid and Fidler, Sanja and Acuna, David},
 doi = {10.48550/arXiv.2404.06510},
 month = {April},
 note = {arXiv:2404.06510},
 publisher = {arXiv},
 title = {Can {Feedback} {Enhance} {Semantic} {Grounding} in {Large} {Vision}-{Language} {Models}?},
 url = {http://arxiv.org/abs/2404.06510},
 urldate = {2025-02-07},
 year = {2024}
}

@misc{2404.06511-min_morevqa:_2024,
 author = {Min, Juhong and Buch, Shyamal and Nagrani, Arsha and Cho, Minsu and Schmid, Cordelia},
 doi = {10.48550/arXiv.2404.06511},
 month = {April},
 note = {arXiv:2404.06511},
 publisher = {arXiv},
 shorttitle = {{MoReVQA}},
 title = {{MoReVQA}: {Exploring} {Modular} {Reasoning} {Models} for {Video} {Question} {Answering}},
 url = {http://arxiv.org/abs/2404.06511},
 urldate = {2025-02-07},
 year = {2024}
}

@misc{2404.07449-ranasinghe_learning_2024,
 author = {Ranasinghe, Kanchana and Shukla, Satya Narayan and Poursaeed, Omid and Ryoo, Michael S. and Lin, Tsung-Yu},
 doi = {10.48550/arXiv.2404.07449},
 month = {April},
 note = {arXiv:2404.07449},
 publisher = {arXiv},
 title = {Learning to {Localize} {Objects} {Improves} {Spatial} {Reasoning} in {Visual}-{LLMs}},
 url = {http://arxiv.org/abs/2404.07449},
 urldate = {2025-02-07},
 year = {2024}
}

@misc{2404.12390-fu_blink:_2024,
 author = {Fu, Xingyu and Hu, Yushi and Li, Bangzheng and Feng, Yu and Wang, Haoyu and Lin, Xudong and Roth, Dan and Smith, Noah A. and Ma, Wei-Chiu and Krishna, Ranjay},
 doi = {10.48550/arXiv.2404.12390},
 month = {July},
 note = {arXiv:2404.12390},
 publisher = {arXiv},
 shorttitle = {{BLINK}},
 title = {{BLINK}: {Multimodal} {Large} {Language} {Models} {Can} {See} but {Not} {Perceive}},
 url = {http://arxiv.org/abs/2404.12390},
 urldate = {2025-02-07},
 year = {2024}
}

@misc{2404.13847-ma_eventlens:_2024,
 author = {Ma, Mingjie and Yu, Zhihuan and Ma, Yichao and Li, Guohui},
 doi = {10.48550/arXiv.2404.13847},
 month = {April},
 note = {arXiv:2404.13847},
 publisher = {arXiv},
 shorttitle = {{EventLens}},
 title = {{EventLens}: {Leveraging} {Event}-{Aware} {Pretraining} and {Cross}-modal {Linking} {Enhances} {Visual} {Commonsense} {Reasoning}},
 url = {http://arxiv.org/abs/2404.13847},
 urldate = {2025-02-07},
 year = {2024}
}

@misc{2404.15190-shin_socratic_2024,
 author = {Shin, Suyeon and jeon, Sujin and Kim, Junghyun and Kang, Gi-Cheon and Zhang, Byoung-Tak},
 doi = {10.48550/arXiv.2404.15190},
 month = {April},
 note = {arXiv:2404.15190},
 publisher = {arXiv},
 shorttitle = {Socratic {Planner}},
 title = {Socratic {Planner}: {Inquiry}-{Based} {Zero}-{Shot} {Planning} for {Embodied} {Instruction} {Following}},
 url = {http://arxiv.org/abs/2404.15190},
 urldate = {2025-02-07},
 year = {2024}
}

@misc{2404.16222-nagarajan_step_2024,
 author = {Nagarajan, Tushar and Torresani, Lorenzo},
 doi = {10.48550/arXiv.2404.16222},
 month = {June},
 note = {arXiv:2404.16222},
 publisher = {arXiv},
 title = {Step {Differences} in {Instructional} {Video}},
 url = {http://arxiv.org/abs/2404.16222},
 urldate = {2025-02-07},
 year = {2024}
}

@misc{2404.18033-huang_exposing_2024,
 author = {Huang, Mingzhen and Jia, Shan and Zhou, Zhou and Ju, Yan and Cai, Jialing and Lyu, Siwei},
 doi = {10.48550/arXiv.2404.18033},
 month = {April},
 note = {arXiv:2404.18033},
 publisher = {arXiv},
 title = {Exposing {Text}-{Image} {Inconsistency} {Using} {Diffusion} {Models}},
 url = {http://arxiv.org/abs/2404.18033},
 urldate = {2025-02-07},
 year = {2024}
}

@misc{2405.03272-zhang_worldqa:_2024,
 author = {Zhang, Yuanhan and Zhang, Kaichen and Li, Bo and Pu, Fanyi and Setiadharma, Christopher Arif and Yang, Jingkang and Liu, Ziwei},
 doi = {10.48550/arXiv.2405.03272},
 month = {May},
 note = {arXiv:2405.03272},
 publisher = {arXiv},
 shorttitle = {{WorldQA}},
 title = {{WorldQA}: {Multimodal} {World} {Knowledge} in {Videos} through {Long}-{Chain} {Reasoning}},
 url = {http://arxiv.org/abs/2405.03272},
 urldate = {2025-02-07},
 year = {2024}
}

@misc{2405.09711-wu_star:_2024,
 author = {Wu, Bo and Yu, Shoubin and Chen, Zhenfang and Tenenbaum, Joshua B. and Gan, Chuang},
 doi = {10.48550/arXiv.2405.09711},
 month = {May},
 note = {arXiv:2405.09711},
 publisher = {arXiv},
 shorttitle = {{STAR}},
 title = {{STAR}: {A} {Benchmark} for {Situated} {Reasoning} in {Real}-{World} {Videos}},
 url = {http://arxiv.org/abs/2405.09711},
 urldate = {2025-02-07},
 year = {2024}
}

@misc{2405.16071-zhao_dynrefer:_2024,
 author = {Zhao, Yuzhong and Liu, Feng and Liu, Yue and Liao, Mingxiang and Gong, Chen and Ye, Qixiang and Wan, Fang},
 doi = {10.48550/arXiv.2405.16071},
 month = {May},
 note = {arXiv:2405.16071},
 publisher = {arXiv},
 shorttitle = {{DynRefer}},
 title = {{DynRefer}: {Delving} into {Region}-level {Multi}-modality {Tasks} via {Dynamic} {Resolution}},
 url = {http://arxiv.org/abs/2405.16071},
 urldate = {2025-02-07},
 year = {2024}
}

@misc{2405.16473-chen_m$^3$cot:_2024,
 author = {Chen, Qiguang and Qin, Libo and Zhang, Jin and Chen, Zhi and Xu, Xiao and Che, Wanxiang},
 doi = {10.48550/arXiv.2405.16473},
 month = {May},
 note = {arXiv:2405.16473},
 publisher = {arXiv},
 shorttitle = {M\${\textasciicircum}3\${CoT}},
 title = {M\${\textasciicircum}3\${CoT}: {A} {Novel} {Benchmark} for {Multi}-{Domain} {Multi}-step {Multi}-modal {Chain}-of-{Thought}},
 url = {http://arxiv.org/abs/2405.16473},
 urldate = {2025-02-07},
 year = {2024}
}

@misc{2405.19209-wang_videotree:_2024,
 author = {Wang, Ziyang and Yu, Shoubin and Stengel-Eskin, Elias and Yoon, Jaehong and Cheng, Feng and Bertasius, Gedas and Bansal, Mohit},
 doi = {10.48550/arXiv.2405.19209},
 month = {October},
 note = {arXiv:2405.19209},
 publisher = {arXiv},
 shorttitle = {{VideoTree}},
 title = {{VideoTree}: {Adaptive} {Tree}-based {Video} {Representation} for {LLM} {Reasoning} on {Long} {Videos}},
 url = {http://arxiv.org/abs/2405.19209},
 urldate = {2025-02-07},
 year = {2024}
}

@misc{2406.00645-fu_furl:_2024,
 author = {Fu, Yuwei and Zhang, Haichao and Wu, Di and Xu, Wei and Boulet, Benoit},
 doi = {10.48550/arXiv.2406.00645},
 month = {June},
 note = {arXiv:2406.00645},
 publisher = {arXiv},
 shorttitle = {{FuRL}},
 title = {{FuRL}: {Visual}-{Language} {Models} as {Fuzzy} {Rewards} for {Reinforcement} {Learning}},
 url = {http://arxiv.org/abs/2406.00645},
 urldate = {2025-02-07},
 year = {2024}
}

@misc{2406.08164-huang_conme:_2024,
 author = {Huang, Irene and Lin, Wei and Mirza, M. Jehanzeb and Hansen, Jacob A. and Doveh, Sivan and Butoi, Victor Ion and Herzig, Roei and Arbelle, Assaf and Kuehne, Hilde and Darrell, Trevor and Gan, Chuang and Oliva, Aude and Feris, Rogerio and Karlinsky, Leonid},
 doi = {10.48550/arXiv.2406.08164},
 month = {November},
 note = {arXiv:2406.08164},
 publisher = {arXiv},
 shorttitle = {{ConMe}},
 title = {{ConMe}: {Rethinking} {Evaluation} of {Compositional} {Reasoning} for {Modern} {VLMs}},
 url = {http://arxiv.org/abs/2406.08164},
 urldate = {2025-02-07},
 year = {2024}
}

@misc{2406.09175-kazemi_remi:_2024,
 author = {Kazemi, Mehran and Dikkala, Nishanth and Anand, Ankit and Devic, Petar and Dasgupta, Ishita and Liu, Fangyu and Fatemi, Bahare and Awasthi, Pranjal and Guo, Dee and Gollapudi, Sreenivas and Qureshi, Ahmed},
 doi = {10.48550/arXiv.2406.09175},
 month = {June},
 note = {arXiv:2406.09175},
 publisher = {arXiv},
 shorttitle = {{ReMI}},
 title = {{ReMI}: {A} {Dataset} for {Reasoning} with {Multiple} {Images}},
 url = {http://arxiv.org/abs/2406.09175},
 urldate = {2025-02-07},
 year = {2024}
}

@misc{2406.09390-reilly_llavidal:_2024,
 author = {Reilly, Dominick and Chakraborty, Rajatsubhra and Sinha, Arkaprava and Govind, Manish Kumar and Wang, Pu and Bremond, Francois and Xue, Le and Das, Srijan},
 doi = {10.48550/arXiv.2406.09390},
 month = {December},
 note = {arXiv:2406.09390},
 publisher = {arXiv},
 shorttitle = {{LLAVIDAL}},
 title = {{LLAVIDAL}: {A} {Large} {LAnguage} {VIsion} {Model} for {Daily} {Activities} of {Living}},
 url = {http://arxiv.org/abs/2406.09390},
 urldate = {2025-02-07},
 year = {2024}
}

@misc{2406.10923-su_investigating_2024,
 author = {Su, Hung-Ting and Chao, Chun-Tong and Hsu, Ya-Ching and Lin, Xudong and Niu, Yulei and Lee, Hung-Yi and Hsu, Winston H.},
 doi = {10.48550/arXiv.2406.10923},
 month = {June},
 note = {arXiv:2406.10923},
 publisher = {arXiv},
 title = {Investigating {Video} {Reasoning} {Capability} of {Large} {Language} {Models} with {Tropes} in {Movies}},
 url = {http://arxiv.org/abs/2406.10923},
 urldate = {2025-02-07},
 year = {2024}
}

@misc{2406.11303-li_videovista:_2024,
 author = {Li, Yunxin and Chen, Xinyu and Hu, Baotian and Wang, Longyue and Shi, Haoyuan and Zhang, Min},
 doi = {10.48550/arXiv.2406.11303},
 month = {June},
 note = {arXiv:2406.11303},
 publisher = {arXiv},
 shorttitle = {{VideoVista}},
 title = {{VideoVista}: {A} {Versatile} {Benchmark} for {Video} {Understanding} and {Reasoning}},
 url = {http://arxiv.org/abs/2406.11303},
 urldate = {2025-02-07},
 year = {2024}
}

@misc{2406.16620-zhang_omagent:_2024,
 author = {Zhang, Lu and Zhao, Tiancheng and Ying, Heting and Ma, Yibo and Lee, Kyusong},
 doi = {10.48550/arXiv.2406.16620},
 month = {November},
 note = {arXiv:2406.16620},
 publisher = {arXiv},
 shorttitle = {{OmAgent}},
 title = {{OmAgent}: {A} {Multi}-modal {Agent} {Framework} for {Complex} {Video} {Understanding} with {Task} {Divide}-and-{Conquer}},
 url = {http://arxiv.org/abs/2406.16620},
 urldate = {2025-02-07},
 year = {2024}
}

@misc{2406.18227-liang_guide:_2024,
 author = {Liang, Jiafeng and Jiang, Shixin and Wang, Zekun and Pan, Haojie and Chen, Zerui and Chu, Zheng and Liu, Ming and Fu, Ruiji and Wang, Zhongyuan and Qin, Bing},
 doi = {10.48550/arXiv.2406.18227},
 month = {June},
 note = {arXiv:2406.18227},
 publisher = {arXiv},
 shorttitle = {{GUIDE}},
 title = {{GUIDE}: {A} {Guideline}-{Guided} {Dataset} for {Instructional} {Video} {Comprehension}},
 url = {http://arxiv.org/abs/2406.18227},
 urldate = {2025-02-07},
 year = {2024}
}

@misc{2406.19392-chen_rextime:_2024,
 author = {Chen, Jr-Jen and Liao, Yu-Chien and Lin, Hsi-Che and Yu, Yu-Chu and Chen, Yen-Chun and Wang, Yu-Chiang Frank},
 doi = {10.48550/arXiv.2406.19392},
 month = {July},
 note = {arXiv:2406.19392},
 publisher = {arXiv},
 shorttitle = {{ReXTime}},
 title = {{ReXTime}: {A} {Benchmark} {Suite} for {Reasoning}-{Across}-{Time} in {Videos}},
 url = {http://arxiv.org/abs/2406.19392},
 urldate = {2025-02-07},
 year = {2024}
}

@misc{2406.19934-cheng_least_2024,
 author = {Cheng, Chuanqi and Guan, Jian and Wu, Wei and Yan, Rui},
 doi = {10.48550/arXiv.2406.19934},
 month = {October},
 note = {arXiv:2406.19934},
 publisher = {arXiv},
 shorttitle = {From the {Least} to the {Most}},
 title = {From the {Least} to the {Most}: {Building} a {Plug}-and-{Play} {Visual} {Reasoner} via {Data} {Synthesis}},
 url = {http://arxiv.org/abs/2406.19934},
 urldate = {2025-02-07},
 year = {2024}
}

@misc{2407.03008-liao_align_2024,
 author = {Liao, Zhaohe and Li, Jiangtong and Niu, Li and Zhang, Liqing},
 doi = {10.48550/arXiv.2407.03008},
 month = {July},
 note = {arXiv:2407.03008},
 publisher = {arXiv},
 shorttitle = {Align and {Aggregate}},
 title = {Align and {Aggregate}: {Compositional} {Reasoning} with {Video} {Alignment} and {Answer} {Aggregation} for {Video} {Question}-{Answering}},
 url = {http://arxiv.org/abs/2407.03008},
 urldate = {2025-02-07},
 year = {2024}
}

@misc{2407.04681-lin_rethinking_2024,
 author = {Lin, Yuanze and Li, Yunsheng and Chen, Dongdong and Xu, Weijian and Clark, Ronald and Torr, Philip and Yuan, Lu},
 doi = {10.48550/arXiv.2407.04681},
 month = {July},
 note = {arXiv:2407.04681},
 publisher = {arXiv},
 title = {Rethinking {Visual} {Prompting} for {Multimodal} {Large} {Language} {Models} with {External} {Knowledge}},
 url = {http://arxiv.org/abs/2407.04681},
 urldate = {2025-02-07},
 year = {2024}
}

@misc{2407.04973-xiao_logicvista:_2024,
 author = {Xiao, Yijia and Sun, Edward and Liu, Tianyu and Wang, Wei},
 doi = {10.48550/arXiv.2407.04973},
 month = {July},
 note = {arXiv:2407.04973},
 publisher = {arXiv},
 shorttitle = {{LogicVista}},
 title = {{LogicVista}: {Multimodal} {LLM} {Logical} {Reasoning} {Benchmark} in {Visual} {Contexts}},
 url = {http://arxiv.org/abs/2407.04973},
 urldate = {2025-02-07},
 year = {2024}
}

@misc{2407.06189-zohar_video-star:_2024,
 author = {Zohar, Orr and Wang, Xiaohan and Bitton, Yonatan and Szpektor, Idan and Yeung-Levy, Serena},
 doi = {10.48550/arXiv.2407.06189},
 month = {July},
 note = {arXiv:2407.06189},
 publisher = {arXiv},
 shorttitle = {Video-{STaR}},
 title = {Video-{STaR}: {Self}-{Training} {Enables} {Video} {Instruction} {Tuning} with {Any} {Supervision}},
 url = {http://arxiv.org/abs/2407.06189},
 urldate = {2025-02-07},
 year = {2024}
}

@misc{2407.11422-zhang_reflective_2024,
 author = {Zhang, Jinrui and Wang, Teng and Zhang, Haigang and Lu, Ping and Zheng, Feng},
 doi = {10.48550/arXiv.2407.11422},
 month = {July},
 note = {arXiv:2407.11422},
 publisher = {arXiv},
 shorttitle = {Reflective {Instruction} {Tuning}},
 title = {Reflective {Instruction} {Tuning}: {Mitigating} {Hallucinations} in {Large} {Vision}-{Language} {Models}},
 url = {http://arxiv.org/abs/2407.11422},
 urldate = {2025-02-07},
 year = {2024}
}

@misc{2407.11522-li_fire:_2024,
 author = {Li, Pengxiang and Gao, Zhi and Zhang, Bofei and Yuan, Tao and Wu, Yuwei and Harandi, Mehrtash and Jia, Yunde and Zhu, Song-Chun and Li, Qing},
 doi = {10.48550/arXiv.2407.11522},
 month = {December},
 note = {arXiv:2407.11522},
 publisher = {arXiv},
 shorttitle = {{FIRE}},
 title = {{FIRE}: {A} {Dataset} for {Feedback} {Integration} and {Refinement} {Evaluation} of {Multimodal} {Models}},
 url = {http://arxiv.org/abs/2407.11522},
 urldate = {2025-02-07},
 year = {2024}
}

@misc{2408.00765-yu_mm-vet_2024,
 author = {Yu, Weihao and Yang, Zhengyuan and Ren, Lingfeng and Li, Linjie and Wang, Jianfeng and Lin, Kevin and Lin, Chung-Ching and Liu, Zicheng and Wang, Lijuan and Wang, Xinchao},
 doi = {10.48550/arXiv.2408.00765},
 month = {December},
 note = {arXiv:2408.00765},
 publisher = {arXiv},
 shorttitle = {{MM}-{Vet} v2},
 title = {{MM}-{Vet} v2: {A} {Challenging} {Benchmark} to {Evaluate} {Large} {Multimodal} {Models} for {Integrated} {Capabilities}},
 url = {http://arxiv.org/abs/2408.00765},
 urldate = {2025-02-07},
 year = {2024}
}

@misc{2408.02032-huo_self-introspective_2024,
 author = {Huo, Fushuo and Xu, Wenchao and Zhang, Zhong and Wang, Haozhao and Chen, Zhicheng and Zhao, Peilin},
 doi = {10.48550/arXiv.2408.02032},
 month = {October},
 note = {arXiv:2408.02032},
 publisher = {arXiv},
 shorttitle = {Self-{Introspective} {Decoding}},
 title = {Self-{Introspective} {Decoding}: {Alleviating} {Hallucinations} for {Large} {Vision}-{Language} {Models}},
 url = {http://arxiv.org/abs/2408.02032},
 urldate = {2025-02-07},
 year = {2024}
}

@misc{2408.02034-huang_mini-monkey:_2024,
 author = {Huang, Mingxin and Liu, Yuliang and Liang, Dingkang and Jin, Lianwen and Bai, Xiang},
 doi = {10.48550/arXiv.2408.02034},
 month = {October},
 note = {arXiv:2408.02034},
 publisher = {arXiv},
 shorttitle = {Mini-{Monkey}},
 title = {Mini-{Monkey}: {Alleviating} the {Semantic} {Sawtooth} {Effect} for {Lightweight} {MLLMs} via {Complementary} {Image} {Pyramid}},
 url = {http://arxiv.org/abs/2408.02034},
 urldate = {2025-02-07},
 year = {2024}
}

@misc{2408.03940-gou_how_2024,
 author = {Gou, Chenhui and Felemban, Abdulwahab and Khan, Faizan Farooq and Zhu, Deyao and Cai, Jianfei and Rezatofighi, Hamid and Elhoseiny, Mohamed},
 doi = {10.48550/arXiv.2408.03940},
 month = {August},
 note = {arXiv:2408.03940},
 publisher = {arXiv},
 title = {How {Well} {Can} {Vision} {Language} {Models} {See} {Image} {Details}?},
 url = {http://arxiv.org/abs/2408.03940},
 urldate = {2025-02-07},
 year = {2024}
}

@misc{2408.05019-wang_instruction_2024,
 author = {Wang, Dongsheng and Cui, Jiequan and Li, Miaoge and Lin, Wang and Chen, Bo and Zhang, Hanwang},
 doi = {10.48550/arXiv.2408.05019},
 month = {August},
 note = {arXiv:2408.05019},
 publisher = {arXiv},
 title = {Instruction {Tuning}-free {Visual} {Token} {Complement} for {Multimodal} {LLMs}},
 url = {http://arxiv.org/abs/2408.05019},
 urldate = {2025-02-07},
 year = {2024}
}

@misc{2408.08632-li_survey_2024,
 author = {Li, Jian and Lu, Weiheng and Fei, Hao and Luo, Meng and Dai, Ming and Xia, Min and Jin, Yizhang and Gan, Zhenye and Qi, Ding and Fu, Chaoyou and Tai, Ying and Yang, Wankou and Wang, Yabiao and Wang, Chengjie},
 doi = {10.48550/arXiv.2408.08632},
 month = {September},
 note = {arXiv:2408.08632},
 publisher = {arXiv},
 title = {A {Survey} on {Benchmarks} of {Multimodal} {Large} {Language} {Models}},
 url = {http://arxiv.org/abs/2408.08632},
 urldate = {2025-02-07},
 year = {2024}
}

@misc{2408.10433-ouali_clip-dpo:_2024,
 author = {Ouali, Yassine and Bulat, Adrian and Martinez, Brais and Tzimiropoulos, Georgios},
 doi = {10.48550/arXiv.2408.10433},
 month = {August},
 note = {arXiv:2408.10433},
 publisher = {arXiv},
 shorttitle = {{CLIP}-{DPO}},
 title = {{CLIP}-{DPO}: {Vision}-{Language} {Models} as a {Source} of {Preference} for {Fixing} {Hallucinations} in {LVLMs}},
 url = {http://arxiv.org/abs/2408.10433},
 urldate = {2025-02-07},
 year = {2024}
}

@misc{2408.11813-yin_sea:_2024,
 author = {Yin, Yuanyang and Zhao, Yaqi and Zhang, Yajie and Lin, Ke and Wang, Jiahao and Tao, Xin and Wan, Pengfei and Zhang, Di and Yin, Baoqun and Zhang, Wentao},
 doi = {10.48550/arXiv.2408.11813},
 month = {August},
 note = {arXiv:2408.11813},
 publisher = {arXiv},
 shorttitle = {{SEA}},
 title = {{SEA}: {Supervised} {Embedding} {Alignment} for {Token}-{Level} {Visual}-{Textual} {Integration} in {MLLMs}},
 url = {http://arxiv.org/abs/2408.11813},
 urldate = {2025-02-07},
 year = {2024}
}

@misc{2408.13890-huang_making_2024,
 author = {Huang, Zhijian and Tang, Tao and Chen, Shaoxiang and Lin, Sihao and Jie, Zequn and Ma, Lin and Wang, Guangrun and Liang, Xiaodan},
 doi = {10.48550/arXiv.2408.13890},
 month = {August},
 note = {arXiv:2408.13890},
 publisher = {arXiv},
 title = {Making {Large} {Language} {Models} {Better} {Planners} with {Reasoning}-{Decision} {Alignment}},
 url = {http://arxiv.org/abs/2408.13890},
 urldate = {2025-02-07},
 year = {2024}
}

@misc{2408.14469-chen_grounded_2024,
 author = {Chen, Qirui and Di, Shangzhe and Xie, Weidi},
 doi = {10.48550/arXiv.2408.14469},
 month = {August},
 note = {arXiv:2408.14469},
 publisher = {arXiv},
 title = {Grounded {Multi}-{Hop} {VideoQA} in {Long}-{Form} {Egocentric} {Videos}},
 url = {http://arxiv.org/abs/2408.14469},
 urldate = {2025-02-07},
 year = {2024}
}

@misc{2408.15556-wang_divide_2024,
 author = {Wang, Wenbin and Ding, Liang and Zeng, Minyan and Zhou, Xiabin and Shen, Li and Luo, Yong and Tao, Dacheng},
 doi = {10.48550/arXiv.2408.15556},
 month = {August},
 note = {arXiv:2408.15556},
 publisher = {arXiv},
 shorttitle = {Divide, {Conquer} and {Combine}},
 title = {Divide, {Conquer} and {Combine}: {A} {Training}-{Free} {Framework} for {High}-{Resolution} {Image} {Perception} in {Multimodal} {Large} {Language} {Models}},
 url = {http://arxiv.org/abs/2408.15556},
 urldate = {2025-02-07},
 year = {2024}
}

@misc{2408.17150-qu_look_2024,
 author = {Qu, Xiaoye and Sun, Jiashuo and Wei, Wei and Cheng, Yu},
 doi = {10.48550/arXiv.2408.17150},
 month = {August},
 note = {arXiv:2408.17150},
 publisher = {arXiv},
 shorttitle = {Look, {Compare}, {Decide}},
 title = {Look, {Compare}, {Decide}: {Alleviating} {Hallucination} in {Large} {Vision}-{Language} {Models} via {Multi}-{View} {Multi}-{Path} {Reasoning}},
 url = {http://arxiv.org/abs/2408.17150},
 urldate = {2025-02-07},
 year = {2024}
}

@misc{2409.08202-hsu_what_2024,
 author = {Hsu, Joy and Mao, Jiayuan and Tenenbaum, Joshua B. and Goodman, Noah D. and Wu, Jiajun},
 doi = {10.48550/arXiv.2409.08202},
 month = {September},
 note = {arXiv:2409.08202},
 publisher = {arXiv},
 title = {What {Makes} a {Maze} {Look} {Like} a {Maze}?},
 url = {http://arxiv.org/abs/2409.08202},
 urldate = {2025-02-07},
 year = {2024}
}

@misc{2409.12953-wang_journeybench:_2025,
 author = {Wang, Zhecan and Liu, Junzhang and Tang, Chia-Wei and Alomari, Hani and Sivakumar, Anushka and Sun, Rui and Li, Wenhao and Atabuzzaman, Md and Ayyubi, Hammad and You, Haoxuan and Ishmam, Alvi and Chang, Kai-Wei and Chang, Shih-Fu and Thomas, Chris},
 doi = {10.48550/arXiv.2409.12953},
 month = {January},
 note = {arXiv:2409.12953},
 publisher = {arXiv},
 shorttitle = {{JourneyBench}},
 title = {{JourneyBench}: {A} {Challenging} {One}-{Stop} {Vision}-{Language} {Understanding} {Benchmark} of {Generated} {Images}},
 url = {http://arxiv.org/abs/2409.12953},
 urldate = {2025-02-07},
 year = {2025}
}

@misc{2409.14750-liu_finecops-ref:_2025,
 author = {Liu, Junzhuo and Yang, Xuzheng and Li, Weiwei and Wang, Peng},
 doi = {10.48550/arXiv.2409.14750},
 month = {January},
 note = {arXiv:2409.14750},
 publisher = {arXiv},
 shorttitle = {{FineCops}-{Ref}},
 title = {{FineCops}-{Ref}: {A} new {Dataset} and {Task} for {Fine}-{Grained} {Compositional} {Referring} {Expression} {Comprehension}},
 url = {http://arxiv.org/abs/2409.14750},
 urldate = {2025-02-07},
 year = {2025}
}

@misc{2409.19339-zhang_visual_2024,
 author = {Zhang, Haowei and Liu, Jianzhe and Han, Zhen and Chen, Shuo and He, Bailan and Tresp, Volker and Xu, Zhiqiang and Gu, Jindong},
 doi = {10.48550/arXiv.2409.19339},
 month = {October},
 note = {arXiv:2409.19339},
 publisher = {arXiv},
 title = {Visual {Question} {Decomposition} on {Multimodal} {Large} {Language} {Models}},
 url = {http://arxiv.org/abs/2409.19339},
 urldate = {2025-02-07},
 year = {2024}
}

@misc{2410.02712-xiong_llava-critic:_2024,
 author = {Xiong, Tianyi and Wang, Xiyao and Guo, Dong and Ye, Qinghao and Fan, Haoqi and Gu, Quanquan and Huang, Heng and Li, Chunyuan},
 doi = {10.48550/arXiv.2410.02712},
 month = {October},
 note = {arXiv:2410.02712},
 publisher = {arXiv},
 shorttitle = {{LLaVA}-{Critic}},
 title = {{LLaVA}-{Critic}: {Learning} to {Evaluate} {Multimodal} {Models}},
 url = {http://arxiv.org/abs/2410.02712},
 urldate = {2025-02-07},
 year = {2024}
}

@misc{2410.02884-zhang_llama-berry:_2024,
 author = {Zhang, Di and Wu, Jianbo and Lei, Jingdi and Che, Tong and Li, Jiatong and Xie, Tong and Huang, Xiaoshui and Zhang, Shufei and Pavone, Marco and Li, Yuqiang and Ouyang, Wanli and Zhou, Dongzhan},
 doi = {10.48550/arXiv.2410.02884},
 month = {November},
 note = {arXiv:2410.02884},
 publisher = {arXiv},
 shorttitle = {{LLaMA}-{Berry}},
 title = {{LLaMA}-{Berry}: {Pairwise} {Optimization} for {O1}-like {Olympiad}-{Level} {Mathematical} {Reasoning}},
 url = {http://arxiv.org/abs/2410.02884},
 urldate = {2025-02-07},
 year = {2024}
}

@misc{2410.03321-ni_visual-o1:_2024,
 author = {Ni, Minheng and Fan, Yutao and Zhang, Lei and Zuo, Wangmeng},
 doi = {10.48550/arXiv.2410.03321},
 month = {October},
 note = {arXiv:2410.03321},
 publisher = {arXiv},
 shorttitle = {Visual-{O1}},
 title = {Visual-{O1}: {Understanding} {Ambiguous} {Instructions} via {Multi}-modal {Multi}-turn {Chain}-of-thoughts {Reasoning}},
 url = {http://arxiv.org/abs/2410.03321},
 urldate = {2025-02-07},
 year = {2024}
}

@misc{2410.03577-zou_look_2024,
 author = {Zou, Xin and Wang, Yizhou and Yan, Yibo and Huang, Sirui and Zheng, Kening and Chen, Junkai and Tang, Chang and Hu, Xuming},
 doi = {10.48550/arXiv.2410.03577},
 month = {October},
 note = {arXiv:2410.03577},
 publisher = {arXiv},
 shorttitle = {Look {Twice} {Before} {You} {Answer}},
 title = {Look {Twice} {Before} {You} {Answer}: {Memory}-{Space} {Visual} {Retracing} for {Hallucination} {Mitigation} in {Multimodal} {Large} {Language} {Models}},
 url = {http://arxiv.org/abs/2410.03577},
 urldate = {2025-02-07},
 year = {2024}
}

@misc{2410.04659-wang_actiview:_2024,
 author = {Wang, Ziyue and Chen, Chi and Luo, Fuwen and Dong, Yurui and Zhang, Yuanchi and Xu, Yuzhuang and Wang, Xiaolong and Li, Peng and Liu, Yang},
 doi = {10.48550/arXiv.2410.04659},
 month = {October},
 note = {arXiv:2410.04659},
 publisher = {arXiv},
 shorttitle = {{ActiView}},
 title = {{ActiView}: {Evaluating} {Active} {Perception} {Ability} for {Multimodal} {Large} {Language} {Models}},
 url = {http://arxiv.org/abs/2410.04659},
 urldate = {2025-02-07},
 year = {2024}
}

@misc{2410.04734-fu_tldr:_2024,
 author = {Fu, Deqing and Xiao, Tong and Wang, Rui and Zhu, Wang and Zhang, Pengchuan and Pang, Guan and Jia, Robin and Chen, Lawrence},
 doi = {10.48550/arXiv.2410.04734},
 month = {October},
 note = {arXiv:2410.04734},
 publisher = {arXiv},
 shorttitle = {{TLDR}},
 title = {{TLDR}: {Token}-{Level} {Detective} {Reward} {Model} for {Large} {Vision} {Language} {Models}},
 url = {http://arxiv.org/abs/2410.04734},
 urldate = {2025-02-07},
 year = {2024}
}

@misc{2410.08209-cao_emerging_2024,
 author = {Cao, Shengcao and Gui, Liang-Yan and Wang, Yu-Xiong},
 doi = {10.48550/arXiv.2410.08209},
 month = {October},
 note = {arXiv:2410.08209},
 publisher = {arXiv},
 title = {Emerging {Pixel} {Grounding} in {Large} {Multimodal} {Models} {Without} {Grounding} {Supervision}},
 url = {http://arxiv.org/abs/2410.08209},
 urldate = {2025-02-07},
 year = {2024}
}

@misc{2410.09421-li_vlfeedback:_2024,
 author = {Li, Lei and Xie, Zhihui and Li, Mukai and Chen, Shunian and Wang, Peiyi and Chen, Liang and Yang, Yazheng and Wang, Benyou and Kong, Lingpeng and Liu, Qi},
 doi = {10.48550/arXiv.2410.09421},
 month = {October},
 note = {arXiv:2410.09421},
 publisher = {arXiv},
 shorttitle = {{VLFeedback}},
 title = {{VLFeedback}: {A} {Large}-{Scale} {AI} {Feedback} {Dataset} for {Large} {Vision}-{Language} {Models} {Alignment}},
 url = {http://arxiv.org/abs/2410.09421},
 urldate = {2025-02-07},
 year = {2024}
}

@misc{2410.09575-wang_reconstructive_2024,
 author = {Wang, Haochen and Zheng, Anlin and Zhao, Yucheng and Wang, Tiancai and Ge, Zheng and Zhang, Xiangyu and Zhang, Zhaoxiang},
 doi = {10.48550/arXiv.2410.09575},
 month = {December},
 note = {arXiv:2410.09575},
 publisher = {arXiv},
 title = {Reconstructive {Visual} {Instruction} {Tuning}},
 url = {http://arxiv.org/abs/2410.09575},
 urldate = {2025-02-07},
 year = {2024}
}

@misc{2410.16198-zhang_improve_2024,
 author = {Zhang, Ruohong and Zhang, Bowen and Li, Yanghao and Zhang, Haotian and Sun, Zhiqing and Gan, Zhe and Yang, Yinfei and Pang, Ruoming and Yang, Yiming},
 doi = {10.48550/arXiv.2410.16198},
 month = {October},
 note = {arXiv:2410.16198},
 publisher = {arXiv},
 title = {Improve {Vision} {Language} {Model} {Chain}-of-thought {Reasoning}},
 url = {http://arxiv.org/abs/2410.16198},
 urldate = {2025-02-07},
 year = {2024}
}

@misc{2410.17885-deng_r-cot:_2024,
 author = {Deng, Linger and Liu, Yuliang and Li, Bohan and Luo, Dongliang and Wu, Liang and Zhang, Chengquan and Lyu, Pengyuan and Zhang, Ziyang and Zhang, Gang and Ding, Errui and Zhu, Yingying and Bai, Xiang},
 doi = {10.48550/arXiv.2410.17885},
 month = {October},
 note = {arXiv:2410.17885},
 publisher = {arXiv},
 shorttitle = {R-{CoT}},
 title = {R-{CoT}: {Reverse} {Chain}-of-{Thought} {Problem} {Generation} for {Geometric} {Reasoning} in {Large} {Multimodal} {Models}},
 url = {http://arxiv.org/abs/2410.17885},
 urldate = {2025-02-07},
 year = {2024}
}

@misc{2410.22315-cascante-bonilla_natural_2024,
 author = {Cascante-Bonilla, Paola and Hou, Yu and Cao, Yang Trista and III, Hal Daumé and Rudinger, Rachel},
 doi = {10.48550/arXiv.2410.22315},
 month = {October},
 note = {arXiv:2410.22315},
 publisher = {arXiv},
 title = {Natural {Language} {Inference} {Improves} {Compositionality} in {Vision}-{Language} {Models}},
 url = {http://arxiv.org/abs/2410.22315},
 urldate = {2025-02-07},
 year = {2024}
}

@misc{2411.00855-cheng_vision-language_2024,
 author = {Cheng, Kanzhi and Li, Yantao and Xu, Fangzhi and Zhang, Jianbing and Zhou, Hao and Liu, Yang},
 doi = {10.48550/arXiv.2411.00855},
 month = {October},
 note = {arXiv:2411.00855},
 publisher = {arXiv},
 title = {Vision-{Language} {Models} {Can} {Self}-{Improve} {Reasoning} via {Reflection}},
 url = {http://arxiv.org/abs/2411.00855},
 urldate = {2025-02-07},
 year = {2024}
}

@misc{2411.10440-xu_llava-cot:_2025,
 author = {Xu, Guowei and Jin, Peng and Li, Hao and Song, Yibing and Sun, Lichao and Yuan, Li},
 doi = {10.48550/arXiv.2411.10440},
 month = {January},
 note = {arXiv:2411.10440},
 publisher = {arXiv},
 shorttitle = {{LLaVA}-{CoT}},
 title = {{LLaVA}-{CoT}: {Let} {Vision} {Language} {Models} {Reason} {Step}-by-{Step}},
 url = {http://arxiv.org/abs/2411.10440},
 urldate = {2025-02-07},
 year = {2025}
}

@misc{2411.12591-zheng_thinking_2024,
 author = {Zheng, Haojie and Xu, Tianyang and Sun, Hanchi and Pu, Shu and Chen, Ruoxi and Sun, Lichao},
 doi = {10.48550/arXiv.2411.12591},
 month = {November},
 note = {arXiv:2411.12591},
 publisher = {arXiv},
 shorttitle = {Thinking {Before} {Looking}},
 title = {Thinking {Before} {Looking}: {Improving} {Multimodal} {LLM} {Reasoning} via {Mitigating} {Visual} {Hallucination}},
 url = {http://arxiv.org/abs/2411.12591},
 urldate = {2025-02-07},
 year = {2024}
}

@misc{2411.17451-li_vlrewardbench:_2024,
 author = {Li, Lei and Wei, Yuancheng and Xie, Zhihui and Yang, Xuqing and Song, Yifan and Wang, Peiyi and An, Chenxin and Liu, Tianyu and Li, Sujian and Lin, Bill Yuchen and Kong, Lingpeng and Liu, Qi},
 doi = {10.48550/arXiv.2411.17451},
 month = {November},
 note = {arXiv:2411.17451},
 publisher = {arXiv},
 shorttitle = {{VLRewardBench}},
 title = {{VLRewardBench}: {A} {Challenging} {Benchmark} for {Vision}-{Language} {Generative} {Reward} {Models}},
 url = {http://arxiv.org/abs/2411.17451},
 urldate = {2025-02-07},
 year = {2024}
}

@misc{2412.02071-xue_progress-aware_2024,
 author = {Xue, Zihui and An, Joungbin and Yang, Xitong and Grauman, Kristen},
 doi = {10.48550/arXiv.2412.02071},
 month = {December},
 note = {arXiv:2412.02071},
 publisher = {arXiv},
 title = {Progress-{Aware} {Video} {Frame} {Captioning}},
 url = {http://arxiv.org/abs/2412.02071},
 urldate = {2025-02-07},
 year = {2024}
}

@misc{2412.02172-wu_visco:_2024,
 author = {Wu, Xueqing and Ding, Yuheng and Li, Bingxuan and Lu, Pan and Yin, Da and Chang, Kai-Wei and Peng, Nanyun},
 doi = {10.48550/arXiv.2412.02172},
 month = {December},
 note = {arXiv:2412.02172},
 publisher = {arXiv},
 shorttitle = {{VISCO}},
 title = {{VISCO}: {Benchmarking} {Fine}-{Grained} {Critique} and {Correction} {Towards} {Self}-{Improvement} in {Visual} {Reasoning}},
 url = {http://arxiv.org/abs/2412.02172},
 urldate = {2025-02-07},
 year = {2024}
}

@misc{2412.03704-wang_scaling_2024,
 author = {Wang, Xiyao and Yang, Zhengyuan and Li, Linjie and Lu, Hongjin and Xu, Yuancheng and Lin, Chung-Ching and Lin, Kevin and Huang, Furong and Wang, Lijuan},
 doi = {10.48550/arXiv.2412.03704},
 month = {December},
 note = {arXiv:2412.03704},
 publisher = {arXiv},
 title = {Scaling {Inference}-{Time} {Search} with {Vision} {Value} {Model} for {Improved} {Visual} {Comprehension}},
 url = {http://arxiv.org/abs/2412.03704},
 urldate = {2025-02-07},
 year = {2024}
}

@misc{2412.04531-zhang_magebench:_2024,
 author = {Zhang, Miaosen and Dai, Qi and Yang, Yifan and Bao, Jianmin and Chen, Dongdong and Qiu, Kai and Luo, Chong and Geng, Xin and Guo, Baining},
 doi = {10.48550/arXiv.2412.04531},
 month = {December},
 note = {arXiv:2412.04531},
 publisher = {arXiv},
 shorttitle = {{MageBench}},
 title = {{MageBench}: {Bridging} {Large} {Multimodal} {Models} to {Agents}},
 url = {http://arxiv.org/abs/2412.04531},
 urldate = {2025-02-07},
 year = {2024}
}

@misc{2412.04903-wang_eaco:_2024,
 author = {Wang, Yongxin and Cao, Meng and Lin, Haokun and Han, Mingfei and Ma, Liang and Jiang, Jin and Cheng, Yuhao and Liang, Xiaodan},
 doi = {10.48550/arXiv.2412.04903},
 month = {December},
 note = {arXiv:2412.04903},
 publisher = {arXiv},
 shorttitle = {{EACO}},
 title = {{EACO}: {Enhancing} {Alignment} in {Multimodal} {LLMs} via {Critical} {Observation}},
 url = {http://arxiv.org/abs/2412.04903},
 urldate = {2025-02-07},
 year = {2024}
}

@misc{2412.05243-chen_compcap:_2024,
 author = {Chen, Xiaohui and Shukla, Satya Narayan and Azab, Mahmoud and Singh, Aashu and Wang, Qifan and Yang, David and Peng, ShengYun and Yu, Hanchao and Yan, Shen and Zhang, Xuewen and He, Baosheng},
 doi = {10.48550/arXiv.2412.05243},
 month = {December},
 note = {arXiv:2412.05243},
 publisher = {arXiv},
 shorttitle = {{CompCap}},
 title = {{CompCap}: {Improving} {Multimodal} {Large} {Language} {Models} with {Composite} {Captions}},
 url = {http://arxiv.org/abs/2412.05243},
 urldate = {2025-02-07},
 year = {2024}
}

@misc{2412.07012-zhang_provision:_2024,
 author = {Zhang, Jieyu and Xue, Le and Song, Linxin and Wang, Jun and Huang, Weikai and Shu, Manli and Yan, An and Ma, Zixian and Niebles, Juan Carlos and Savarese, Silvio and Xiong, Caiming and Chen, Zeyuan and Krishna, Ranjay and Xu, Ran},
 doi = {10.48550/arXiv.2412.07012},
 month = {December},
 note = {arXiv:2412.07012},
 publisher = {arXiv},
 shorttitle = {{ProVision}},
 title = {{ProVision}: {Programmatically} {Scaling} {Vision}-centric {Instruction} {Data} for {Multimodal} {Language} {Models}},
 url = {http://arxiv.org/abs/2412.07012},
 urldate = {2025-02-07},
 year = {2024}
}

@misc{2412.08564-shlapentokh-rothman_can_2024,
 author = {Shlapentokh-Rothman, Michal and Wang, Yu-Xiong and Hoiem, Derek},
 doi = {10.48550/arXiv.2412.08564},
 month = {December},
 note = {arXiv:2412.08564},
 publisher = {arXiv},
 title = {Can {We} {Generate} {Visual} {Programs} {Without} {Prompting} {LLMs}?},
 url = {http://arxiv.org/abs/2412.08564},
 urldate = {2025-02-07},
 year = {2024}
}

@misc{2412.08635-sun_multimodal_2024,
 author = {Sun, Yutao and Bao, Hangbo and Wang, Wenhui and Peng, Zhiliang and Dong, Li and Huang, Shaohan and Wang, Jianyong and Wei, Furu},
 doi = {10.48550/arXiv.2412.08635},
 month = {December},
 note = {arXiv:2412.08635},
 publisher = {arXiv},
 title = {Multimodal {Latent} {Language} {Modeling} with {Next}-{Token} {Diffusion}},
 url = {http://arxiv.org/abs/2412.08635},
 urldate = {2025-02-07},
 year = {2024}
}
\bibliographystyle{colm2025_conference}

\clearpage
\appendix

\section{Implications from the Learning Science Perspective}
The design of CoT reasoning for LLMs aligns with several key cognitive science and learning science theories that enhance structured reasoning, problem-solving, and knowledge construction. 

One key theory is Cognitive Load Theory~\citep{sweller2011cognitive,paas2004cognitive}, which reduces a human's working memory workload by breaking complex problems into manageable steps. 
CoT reasoning follows this approach by prompting models to generate step-by-step solutions rather than arriving at an answer in one leap. This structured decomposition mirrors how human learners handle intricate problems by offloading cognitive effort across multiple processing stages. 
Similarly, in the context of learning, scaffolding refers to offering temporary assistance to learners, enabling them to accomplish tasks that would be difficult or impossible for them to complete independently. This support is gradually removed as learners develop the skills and confidence required to perform the task on their own~\citep{van2010scaffolding}.
This supports CoT’s design by guiding LLMs through intermediate steps, much like how educators provide structured support to help students grasp new concepts before transitioning to independent problem-solving.
These theories are both rooted in the Zone of Proximal Development (ZPD), a key construct in Lev Vygotsky's theory of learning and development~\citep{vygotsky1978mind}.

CoT reasoning also aligns with metacognition~\citep{lai2011metacognition} and self-explanation~\citep{bisra2018inducing,chi1994eliciting} by encouraging models to ``think about their thinking''. 
Just as self-explanation improves human learning by prompting individuals to articulate their reasoning~\citep{chi1994eliciting,vanlehn1992model}, CoT forces LLMs to justify their steps, reducing reliance on shallow heuristics. 
This ties into Dual-Process Theory~\citep{frankish2010dual}, where CoT shifts LLMs from fast, intuitive decision-making (System 1) to deliberate, analytical reasoning (System 2), leading to more logical and consistent outputs.

Moreover, CoT fosters constructivist learning~\citep{waite2022jean,bruner2009process} by enabling LLMs to incrementally build knowledge structures through active reasoning. Instead of passively retrieving answers from training data, the model synthesizes prior knowledge with new information, improving adaptability. 
Additionally, analogical reasoning~\citep{gentner2017analogical} plays a role in CoT by helping LLMs map relationships between concepts, allowing them to generalize problem-solving strategies across different contexts.

By integrating these cognitive science and learning science principles, CoT reasoning enhances the interpretability, reliability, and generalization of LLM outputs, making them more aligned with human cognitive processes and educational best practices.

\tikzstyle{my-box}=[
    rectangle,
    draw=black,
    rounded corners,
    text opacity=1,
    minimum height=2em,
    minimum width=5em,
    inner sep=2pt,
    align=center,
    fill opacity=.5,
    line width=0.8pt,
]
\tikzstyle{leaf}=[my-box, minimum height=2em,
    fill=white!80, text=black, align=left,font=\normalsize,
    inner xsep=2pt,
    inner ysep=4pt,
    line width=0.8pt,
]

\begin{figure*}[!ht]
    \centering
    \resizebox{0.85\textwidth}{!}{
        \begin{forest}
            forked edges,
            for tree={
                grow=east,
                reversed=true,
                anchor=base west,
                parent anchor=east,
                child anchor=west,
                base=left,
                font=\large,
                rectangle,
                draw=black,
                rounded corners,
                align=left,
                minimum width=6em,
                edge+={darkgray, line width=1pt},
                s sep=20pt,
                inner xsep=3pt,
                inner ysep=3pt,
                line width=0.8pt,
                ver/.style={rotate=90, child anchor=north, parent anchor=south, anchor=center},
            },
            where level=1{text width=7.0em,font=\normalsize,}{},
            where level=2{text width=10em,font=\normalsize,}{},
            where level=3{text width=8.2em,font=\normalsize,}{},
            [
                \textbf{Inference}
                [
                    \textbf{Adaptive} \\ \textbf{Inference}
                    [
                        ~\cite{2501.13926-guo_can_2025}{, }
                        ~\cite{2501.02964-hu_socratic_2025}{, }
                        ~\cite{2412.10471-yang_vca:_2024}{, }\\
                        ~\cite{2412.02071-xue_progress-aware_2024}{, }
                        ~\cite{2412.02172-wu_visco:_2024}{, }
                        ~\cite{2412.03704-wang_scaling_2024}{, }\\
                        ~\cite{2410.03577-zou_look_2024}{, }
                        ~\cite{2408.17150-qu_look_2024}{, }
                        ~\cite{2410.16198-zhang_improve_2024}{, }\\
                        ~\cite{2408.02032-huo_self-introspective_2024}{, }
                        ~\cite{2407.03008-liao_align_2024}{, }
                        ~\cite{2406.10923-su_investigating_2024}{, }\\
                        ~\cite{2406.11303-li_videovista:_2024}{, }
                        ~\cite{2408.15556-wang_divide_2024}{, }
                        ~\cite{2411.10440-xu_llava-cot:_2025}{, }\\
                        ~\cite{2405.16071-zhao_dynrefer:_2024}{, }
                        ~\cite{2404.06510-liao_can_2024}{, }
                        ~\cite{2403.14743-mahmood_vurf:_2024}{, }\\
                        ~\cite{2402.14767-cao_dualfocus:_2024}{, }
                        ~\cite{2405.09711-wu_star:_2024}{, }
                        ~\cite{2312.17240-yang_lisa++:_2024}{, }\\
                        ~\cite{2311.13601-li_visual_2023}{, }
                        ~\cite{2311.06612-pi_perceptiongpt:_2023}
                        , leaf, text width=30em
                    ]
                ]
                [
                    \textbf{Reward} \\ \textbf{Models}
                    [
                        ~\cite{2501.13926-guo_can_2025}{, }
                        ~\cite{2501.05444-hao_can_2025}{, }
                        ~\cite{2411.17451-li_vlrewardbench:_2024}{, }\\
                        ~\cite{2406.00645-fu_furl:_2024}{, }
                        ~\cite{2403.12884-ke_hydra:_2024}{, }
                        ~\cite{2410.02712-xiong_llava-critic:_2024}{, }\\
                        ~\cite{2410.22315-cascante-bonilla_natural_2024}{, }
                        ~\cite{2411.00855-cheng_vision-language_2024}{, }\\
                        ~\cite{2410.16198-zhang_improve_2024}{, }
                        , leaf, text width=30em
                    ]
                ]
                [
                    \textbf{Feedback}
                    [
                        ~\cite{2501.13926-guo_can_2025}{, }
                        ~\cite{2412.02172-wu_visco:_2024}{, }
                        ~\cite{2501.02964-hu_socratic_2025}{, }\\
                        ~\cite{2503.07365-meng_mm-eureka:_2025}{, }
                        ~\cite{2407.11522-li_fire:_2024}{, }
                        ~\cite{2410.02712-xiong_llava-critic:_2024}{, }\\
                        ~\cite{2411.00855-cheng_vision-language_2024}{, }
                        ~\cite{2410.16198-zhang_improve_2024}{, }
                        ~\cite{2411.12591-zheng_thinking_2024}{, }\\
                        ~\cite{2404.06510-liao_can_2024}{, }
                        ~\cite{2402.01345-han_skip_2024}{, }
                        ~\cite{2405.19209-wang_videotree:_2024}{, }\\
                        ~\cite{2403.16999-shao_visual_2024}{, }
                        ~\cite{2404.04007-liang_neural-symbolic_2024}{, }\\
                        ~\cite{2311.07362-lee_volcano:_2024}{, }
                        ~\cite{2312.03052-hu_visual_2024}{, }
                        , leaf, text width=30em
                    ]
                ]
                [
                    \textbf{Multimodal} \\ \textbf{Decomposition}
                    [
                        ~\cite{2503.06252-xiang_can_2025}{, }
                        ~\cite{2501.13536-liang_reasvqa:_2025}{, }
                        ~\cite{2501.05069-liu_commonsense_2025}{, }\\
                        ~\cite{2412.08564-shlapentokh-rothman_can_2024}{, }
                        ~\cite{2409.19339-zhang_visual_2024}{, }\\
                        ~\cite{2408.14469-chen_grounded_2024}{, }
                        ~\cite{2409.08202-hsu_what_2024}{, }
                        ~\cite{2406.10923-su_investigating_2024}{, }\\
                        ~\cite{2406.09390-reilly_llavidal:_2024}{, }
                        ~\cite{2408.05019-wang_instruction_2024}{, }
                        ~\cite{2406.16620-zhang_omagent:_2024}{, }\\
                        ~\cite{2404.06511-min_morevqa:_2024}{, }
                        ~\cite{2404.01299-parmar_causalchaos!_2024}{, }\\
                        ~\cite{2403.09394-wang_git:_2024}{, }
                        ~\cite{2404.01911-dzabraev_vlrm:_2024}{, }
                        ~\cite{2405.09711-wu_star:_2024}{, }\\
                        ~\cite{2312.14135-wu_v*:_2023}{, }
                        ~\cite{2309.15112-zhang_internlm-xcomposer:_2023}{, }
                        , leaf, text width=30em
                    ]
                ]
                [
                    \textbf{Efficiency and} \\ \textbf{Scalability}
                    [
                        ~\cite{2502.00382-goyal_masked_2025}{, }
                        ~\cite{2501.07542-li_imagine_2025}{, }
                        ~\cite{2412.08635-sun_multimodal_2024}{, }\\
                        ~\cite{2412.18072-fan_mmfactory:_2024}{, }
                        ~\cite{2501.02669-park_generalizing_2025}{, }
                        ~\cite{2408.15556-wang_divide_2024}{, }\\
                        ~\cite{2409.12953-wang_journeybench:_2025}{, }
                        ~\cite{2408.02032-huo_self-introspective_2024}{, }\\
                        ~\cite{2407.04681-lin_rethinking_2024}{, }
                        ~\cite{2410.03321-ni_visual-o1:_2024}{, }
                        ~\cite{2405.19209-wang_videotree:_2024}{, }\\
                        ~\cite{2406.08164-huang_conme:_2024}{, }
                        ~\cite{2403.07487-zhang_motion_2024}{, }
                        ~\cite{2403.09333-zhan_griffon_2024}{, }\\
                        ~\cite{2403.19322-chen_plug-and-play_2024}{, }
                        ~\cite{2403.09394-wang_git:_2024}{, }
                        ~\cite{2312.11420-zhao_tuning_2023}{, }\\
                        ~\cite{2311.06612-pi_perceptiongpt:_2023}{, }
                        ~\cite{2401.12168-chen_spatialvlm:_2024}{  }
                        , leaf, text width=30em
                    ]
                ]
                [
                    \textbf{Search} \\ \textbf{Strategies}
                    [
                        ~\cite{2412.18319-yao_mulberry:_2024}{, }
                        ~\cite{2412.10471-yang_vca:_2024}{, }
                        ~\cite{2412.03704-wang_scaling_2024}{, }\\
                        ~\cite{2501.10674-imam_can_2025}{, }
                        ~\cite{2411.10440-xu_llava-cot:_2025}{, }
                        ~\cite{2408.17150-qu_look_2024}{, }\\
                        ~\cite{2408.15556-wang_divide_2024}{, }
                        ~\cite{2406.11303-li_videovista:_2024}{, }
                        ~\cite{2406.09175-kazemi_remi:_2024}{, }\\
                        ~\cite{2404.15190-shin_socratic_2024}{, }
                        ~\cite{2405.03272-zhang_worldqa:_2024}{, }\\
                        ~\cite{2405.16071-zhao_dynrefer:_2024}{, }
                        ~\cite{2406.00645-fu_furl:_2024}{, }
                        ~\cite{2403.14743-mahmood_vurf:_2024}{, }\\
                        ~\cite{2312.14135-wu_v*:_2023}{, }
                        ~\cite{2306.08129-hu_avis:_2023}{, }
                        , leaf, text width=30em
                    ]
                ]
            ]
        \end{forest}
    }
    \caption{Comprehensive Overview of Methods and Frameworks focus on test-time compute}
    \label{fig:taxonomy}
\end{figure*}
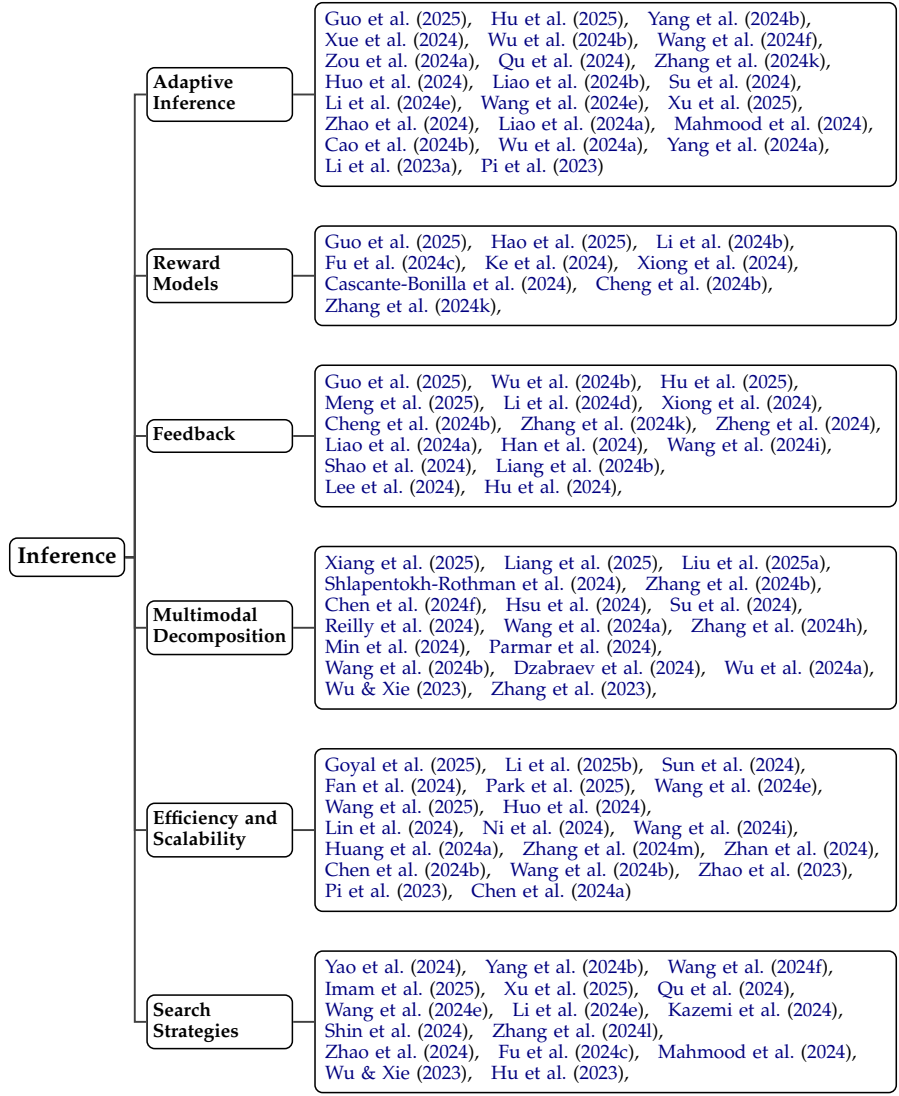

\begin{figure*}[!ht]
    \centering
    \resizebox{0.85\textwidth}{!}{
        \begin{forest}
            forked edges,
            for tree={
                grow=east,
                reversed=true,
                anchor=base west,
                parent anchor=east,
                child anchor=west,
                base=left,
                font=\large,
                rectangle,
                draw=black,
                rounded corners,
                align=left,
                minimum width=6em,
                edge+={darkgray, line width=1pt},
                s sep=20pt,
                inner xsep=3pt,
                inner ysep=3pt,
                line width=0.8pt,
                ver/.style={rotate=90, child anchor=north, parent anchor=south, anchor=center},
            },
            where level=1{text width=7.0em,font=\normalsize,}{},
            where level=2{text width=10em,font=\normalsize,}{},
            where level=3{text width=8.2em,font=\normalsize,}{},
            [
                \textbf{Post} \\ \textbf{Training}
                [
                    \textbf{Policy} \\ \textbf{Optimization}
                    [
                        ~\cite{2406.00645-fu_furl:_2024}{, }
                        ~\cite{2402.07384-zhang_exploring_2024}{, }
                        ~\cite{2312.03052-hu_visual_2024}{, }\\
                        ~\cite{2310.07932-tian_what_2024}{, }
                        ~\cite{2309.10790-kim_guide_2023}{, }
                        ~\cite{2402.06118-yan_vigor:_2024}{, }\\
                        ~\cite{2408.10433-ouali_clip-dpo:_2024}{, }
                        ~\cite{2410.02712-xiong_llava-critic:_2024}{, }
                        ~\cite{2410.04659-wang_actiview:_2024}{, }\\
                        ~\cite{2501.18880-waite_rls3:_2025}{, }
                        ~\cite{2503.08525-wei_gtr:_2025}{, }
                        ~\cite{2503.05379-zhao_r1-omni:_2025}{, }\\
                        ~\cite{2411.10440-xu_llava-cot:_2025}{, }
                        ~\cite{2410.17885-deng_r-cot:_2024}{, }
                        , leaf, text width=30em
                    ]
                ]
                [
                    \textbf{Model} \\ \textbf{Architecture}
                    [
                        ~\cite{2406.09390-reilly_llavidal:_2024}{, }
                        ~\cite{2404.13847-ma_eventlens:_2024}{, }
                        ~\cite{2404.01258-zhang_direct_2024}{, }\\
                        ~\cite{2403.09333-zhan_griffon_2024}{, }
                        ~\cite{2401.17981-jiao_training-free_2024}{, }\\
                        ~\cite{2401.03105-he_incorporating_2024}{, }
                        ~\cite{2312.08870-ma_vista-llama:_2023}{, }
                        ~\cite{2312.03631-ben-kish_mitigating_2024}{, }\\
                        ~\cite{2312.03052-hu_visual_2024}{, }
                        ~\cite{2311.03079-wang_cogvlm:_2024}{, }
                        ~\cite{2410.09575-wang_reconstructive_2024}{, }\\
                        ~\cite{2410.08209-cao_emerging_2024}{, }
                        ~\cite{2408.14469-chen_grounded_2024}{, }
                        ~\cite{2408.13890-huang_making_2024}{, }
                        , leaf, text width=30em
                    ]
                ]
                [
                    \textbf{Reward} \\ \textbf{Alignment}
                    [
                        ~\cite{2501.05444-hao_can_2025}{, }
                        ~\cite{2411.17451-li_vlrewardbench:_2024}{, }
                        ~\cite{2410.02712-xiong_llava-critic:_2024}{, }\\
                        ~\cite{2403.12884-ke_hydra:_2024}{, }
                        ~\cite{2406.00645-fu_furl:_2024}{, }
                        ~\cite{2404.01911-dzabraev_vlrm:_2024}{, }\\
                        ~\cite{2312.03052-hu_visual_2024}{, }
                        , leaf, text width=30em
                    ]
                ]
                [
                    \textbf{Iterative} \\ \textbf{Refinement}
                    [
                        ~\cite{2503.07365-meng_mm-eureka:_2025}{, }
                        ~\cite{2411.12591-zheng_thinking_2024}{, }
                        ~\cite{2404.06510-liao_can_2024}{, }\\
                        ~\cite{2412.02172-wu_visco:_2024}{, }
                        ~\cite{2403.16999-shao_visual_2024}{, }\\
                        ~\cite{2402.01345-han_skip_2024}{, }
                        ~\cite{2312.03052-hu_visual_2024}{, }
                        ~\cite{2311.07362-lee_volcano:_2024}{, }\\
                        , leaf, text width=30em
                    ]
                ]
                [
                    \textbf{Spatial} \\ \textbf{Temporal} \\ \textbf{Modeling}
                    [
                        ~\cite{2405.19209-wang_videotree:_2024}{, }
                        ~\cite{2403.07487-zhang_motion_2024}{, }
                        ~\cite{2401.12168-chen_spatialvlm:_2024}{, }\\
                        ~\cite{2312.14135-wu_v*:_2023}{, }
                        , leaf, text width=30em
                    ]
                ]
                [
                    \textbf{Dataset} \\ \textbf{Curation}
                    [
                        ~\cite{2412.18072-fan_mmfactory:_2024}{, }
                        ~\cite{2409.12953-wang_journeybench:_2025}{, }\\
                        ~\cite{2404.16222-nagarajan_step_2024}{, }
                        ~\cite{2405.03272-zhang_worldqa:_2024}{, }\\
                        ~\cite{2407.04681-lin_rethinking_2024}{, }
                        ~\cite{2312.11420-zhao_tuning_2023}{, }
                        ~\cite{2306.08129-hu_avis:_2023}{, }
                        , leaf, text width=30em
                    ]
                ]
            ]
        \end{forest}
    }
    \caption{Comprehensive Overview of Methods and Frameworks focus on post-training improvement}
    \label{fig:taxonomy}
\end{figure*}
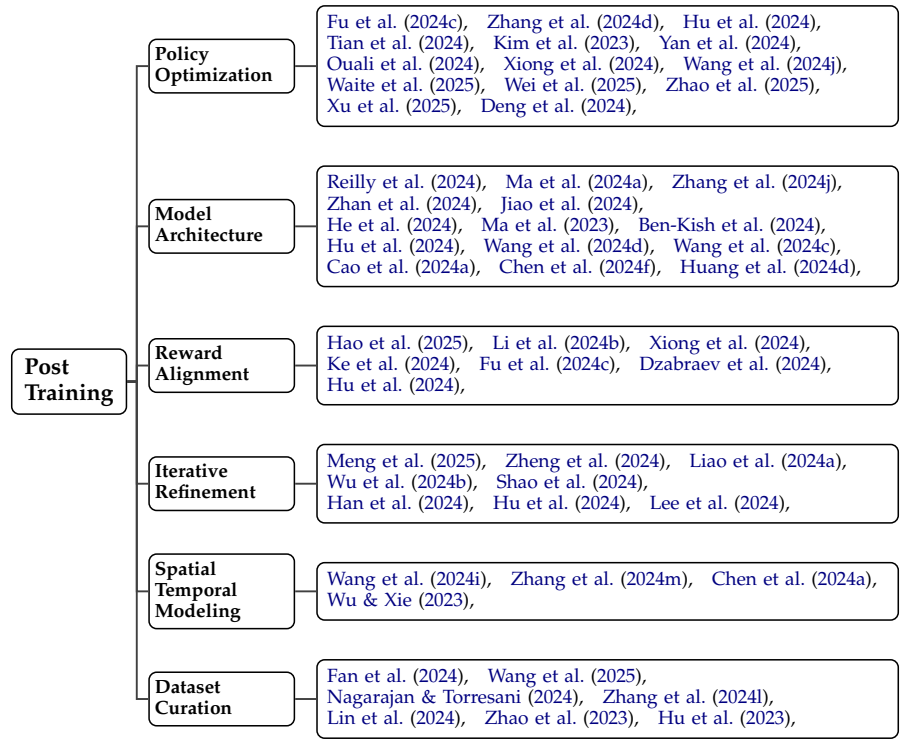

\end{document}